\documentclass[10pt,journal,compsoc]{IEEEtran}
\ifCLASSOPTIONcompsoc
 \usepackage{cite}
\else
 \usepackage{cite}
\fi
\usepackage{xspace}
\usepackage{tabularx, booktabs}

\ifCLASSINFOpdf
 \usepackage[pdftex]{graphicx}
 \graphicspath{{Images/}}
 \DeclareGraphicsExtensions{.pdf,.jpeg,.png}
\else
\fi
\usepackage{booktabs}

\usepackage{algorithmic}

\usepackage{xcolor}

\ifCLASSOPTIONcompsoc
 \usepackage[caption=false,font=footnotesize,labelfont=sf,textfont=sf]{subfig}
 
\else
 \usepackage[caption=false,font=footnotesize]{subfig}
\fi
\usepackage{caption}
\usepackage{color}
\usepackage[algo2e,ruled,vlined]{algorithm2e}
\usepackage{amsmath}
\usepackage{amssymb}
\usepackage{hyperref}
\usepackage{xcolor}
\usepackage[]{algorithm}
\usepackage{comment}
\usepackage{multirow}

\newcommand*{\colorboxed}{}
\def\colorboxed#1#{%
 \colorboxedAux{#1}%
}
\newcommand*{\colorboxedAux}[3]{%
 \begingroup
 \colorlet{cb@saved}{.}%
 \color#1{#2}%
 \boxed{%
 \color{cb@saved}%
 #3%
 }%
 \endgroup
}

\usepackage{array}
\newcolumntype{L}[1]{>{\raggedright\let\newline\\\arraybackslash\hspace{0pt}}m{#1}}
\newcolumntype{C}[1]{>{\centering\let\newline\\\arraybackslash\hspace{0pt}}m{#1}}
\newcolumntype{R}[1]{>{\raggedleft\let\newline\\\arraybackslash\hspace{0pt}}m{#1}}

\newcommand{\beq}{\begin{equation}}
\newcommand{\eeq}{\end{equation}}

\renewcommand{\vec}[1]{\mathbf{#1}}

\newcommand{\mr}[1]{\mathrm{#1}}

\newcommand{\xx}{\vec{x}}

\begin{document}
%

\title{IVD-Net: Intervertebral disc localization and segmentation in MRI with a multi-modal UNet}
%
%
%
%

\author{Jose~Dolz,
 Christian~Desrosiers,
 and~Ismail~Ben~Ayed
\IEEEcompsocitemizethanks{\IEEEcompsocthanksitem J. Dolz, C. Desrosiers and I. Ben Ayed are with the ETS Montreal, Canada. email:[jose.dolz; christian.desrosiers;ismail.benayed]@etsmtl.ca}

\thanks{Manuscript submitted to the Proceedings of the MICCAI 2018 IVD Challenge}}

\IEEEtitleabstractindextext{%
\begin{abstract}

Accurate localization and segmentation of intervertebral disc (IVD) is crucial for the assessment of spine disease diagnosis. Despite the technological advances in medical imaging, IVD localization and segmentation are still manually performed, which is time-consuming and prone to errors. If, in addition, multi-modal imaging is considered, the burden imposed on disease assessments increases substantially. In this paper, we propose an architecture for IVD localization and segmentation in multi-modal magnetic resonance images (MRI), which extends the well-known UNet. Compared to single images, multi-modal data brings complementary information, contributing to better data representation and discriminative power. Our contributions are three-fold. First, how to effectively integrate and fully leverage multi-modal data remains almost unexplored. In this work, each MRI modality is processed in a different path to better exploit their unique information. Second, inspired by HyperDenseNet \cite{dolz2018hyperdense}, the network is densely-connected both within each path and across different paths, granting the model the freedom to learn \textit{where} and \textit{how} the different modalities should be processed and combined. Third, we improved standard U-Net modules by extending inception modules \cite{szegedy2016rethinking} with two convolutional blocks with dilated convolutions of different scale, which helps handling multi-scale context. We report experiments over the data set of the public MICCAI 2018 Challenge on Automatic Intervertebral Disc Localization and Segmentation, with 13 multi-modal MRI images used for training and 3 for validation. 
We trained IVD-Net on an NVidia TITAN XP GPU with 16 GBs RAM, using ADAM as optimizer and a learning rate of 1$\times$10$^{-5}$ during 200 epochs. Training took about 5 hours, and segmentation of a whole volume about 2-3 seconds, on average. Several baselines, with different multi-modal fusion strategies, were used to demonstrate the effectiveness of the proposed architecture.
\end{abstract}

\begin{IEEEkeywords}
Deep learning, Intervertebral disc segmentation, medical image segmentation, multi-modal imaging, densely connected network
\end{IEEEkeywords}}

\maketitle

\IEEEdisplaynontitleabstractindextext

\IEEEpeerreviewmaketitle

\IEEEraisesectionheading{\section{Introduction}\label{sec:introduction}}

%
%
%
%

\begin{figure*}[ht]
\centering
\captionsetup{justification=centering}
\includegraphics[width=1\textwidth]{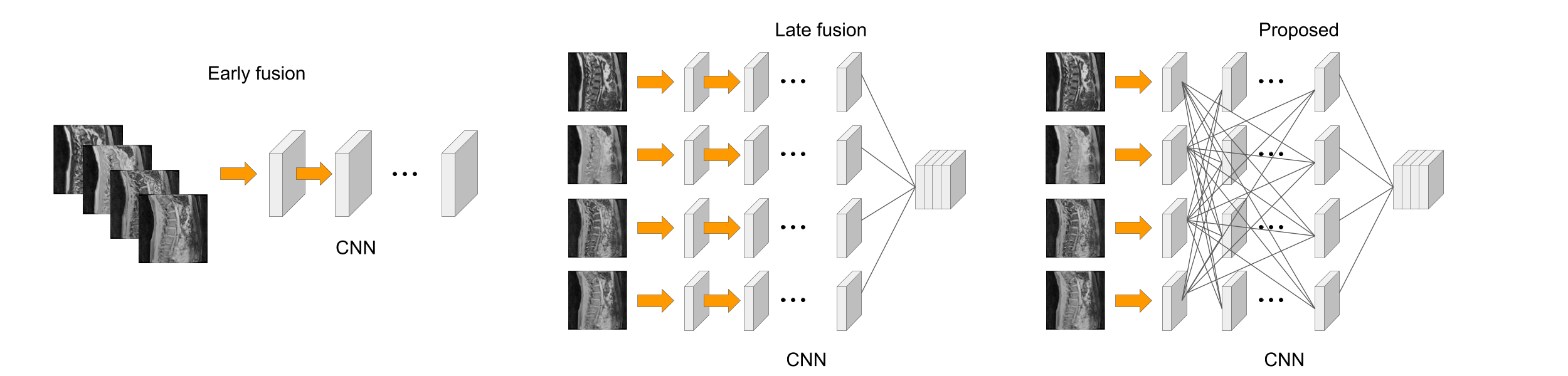}
\caption{Typical features fusion strategies (\textit{left} and \textit{middle}) and proposed fusion technique \textit{right}.} 
\label{fig:fusionStrat}
\end{figure*}

Intervertebral disc (IVD) degeneration \cite{an2004introduction} is one of the main causes for chronic low back pain (LBP), which has become a major public health problem in our society and a leading cause of function incapacity \cite{wieser2011cost}. Magnetic resonance imaging (MRI) is the preferred modality to evaluate lumbar degenerative disc disease  because it offers a good soft tissue contrast without ionizing radiation \cite{hamanishi1994cross}. Advances in multi-modal MRI have increased the quality of diagnosis, treatment and follow-up in many diseases. However this comes at the cost of an increased amount of data, imposing a burden on disease assessments. Visual inspections of such an enormous amount of medical images are prohibitively time-consuming, prone to errors and unsuitable for large-scale studies. 
Developing robust methods for automatic IVD localization and segmentation from multi-modal MRI is thus essential for the diagnosis and treatment of spine pathologies. Having such methods could also reduce the manual work required by clinicians, and provide a faster and more consistent diagnosis.

 
Over the years, various semi-automated and automated techniques have been proposed for IVD localization and segmentation \cite{ayed2011graph,chen2015localization}. Recently, deep convolutional neural networks (CNNs) have shown outstanding performance for this task, outperforming previous segmentation approaches \cite{chen20163d,ji2016automated,kim2018fine,zeng2017dsms,zheng2017evaluation}. For example, Ji et al. \cite{ji2016automated} proposed a standard CNN for IVD segmentation, where the inference was performed pixel-wise by extracting a patch around each pixel. In addition, the authors evaluated different patch strategies, such as 2D or 2.5D patches, as well as the impact of vicinity size. More recently, a deeply supervised multi-scale fully CNN was proposed in \cite{zeng2017dsms} for the segmentation of IVDs in MR-T2 weighted images. An interesting feature of this work is its use of multi-scale deep supervision in the architecture, which alleviates the risk of vanishing gradient during training. Despite achieving satisfactory results, these works have mostly focused on single-modality scenarios.
 
Integrating multi-modal images in deep learning segmentation methods has also gained growing attention recently. Multi-modal segmentation in CNNs is typically addressed with an \emph{early fusion} strategy, where multiple modalities are merged from the original input space of low-level features \cite{dolz2017deep,kamnitsas2017efficient,moeskops2016automatic,valverde2017improving,zhang2015deep} (See Fig. \ref{fig:fusionStrat}, \textit{left}). By concatenating image modalities at the input of the network, we explicitly assume that the relation between different modalities is simple (e.g., linear), which may not correspond to the characteristics of the multi-modal data at hand \cite{Srivastava14}. To better account for the complexity of multi-modal data, other studies investigated \emph{late fusion} strategies \cite{nie2016fully}, where each modality is processed by an independent CNN and the multi-modal outputs are merged in a deep layer, as in the architecture depicted in Fig. \ref{fig:fusionStrat}, \textit{middle}. This \emph{late fusion} strategy was demonstrated to outperform \emph{early fusion} on infant brain segmentation \cite{nie2016fully}. More recently, Aig\"{u}n et al. explored different ways of combining multiple modalities \cite{aygun2018multi}. In this work, all modalities are considered as separate inputs to different CNNs, which are later fused at an `early', `middle' or `late' point. Although it was found that `late' fusion provides better performance, as in \cite{nie2016fully}, this method relies on a single-layer fusion to model the relation between all modalities. Nevertheless, as demonstrated in several works \cite{Srivastava14}, relations between different modalities may be highly complex and cannot easily be modeled by a single layer. To account for the non-linearity in multi-modal data modeling, we recently proposed a CNN that incorporates dense connections not only between pairs of layers within the same path, but also between layers across different paths \cite{dolz2018isointense,dolz2018hyperdense}. This architecture, known as {\em HyperDenseNet}, obtained very competitive performance in the context of infant and adult brain tissue segmentation with multi-modal MRI data.

In the context of IVD localization and segmentation, Li et al. \cite{li20183d} have also considered multi-modal images. Specifically, they proposed a multi-scale and modality dropout learning framework, which employed four MRI modalities. To capture multi-scale context and handle the scale variations of IVDs, three different paths process regions extracted from the same location but at different scales. In addition, a random modality voxel dropout strategy is used to reduce feature co-adaptation between multiple modalities, and encourage each single modality to learn discriminative information independently.

Nevertheless, the combination of multi-modal data at various levels of abstraction has not been fully exploited for IVD localization and segmentation. In this work, we adopt the strategy presented in \cite{dolz2018isointense,dolz2018hyperdense} and propose a multi-path architecture \cite{dolz2018dense} called IVD-Net, where each modality is employed as input of one pathway, with dense connectivity used between the layers, within and across paths (Fig. \ref{fig:fusionStrat}, \textit{right}). Furthermore, we extend the standard convolutional module of InceptionNet \cite{szegedy2016rethinking} by including two additional dilated convolutional blocks, which can help to learn larger context. In our previous work on multi-modal ischemic stroke lesion segmentation \cite{dolz2018dense}, we showed this model to outperform architectures based on early and late fusion, as well as several state-of-art segmentation networks.

\section{Methodology}

The proposed IVD-Net architecture follow the structure of UNet \cite{ronneberger2015u}. This well-known model is composed of two paths: one contracting and one expanding. While the former collapses the input image into a set of high level features forming a compact intermediate representation of the input, the latter employs these features to generate a pixel-wise segmentation mask. Furthermore, it includes skip-connections, which connect the outputs from shallow layers to the input of subsequent layers, with the goal of transferring information that may have been lost in the encoding path during the compression process. 

\subsection{Processing multiple modalities separately}

\begin{figure*}[h!]
\centering
\includegraphics[width=1\textwidth]{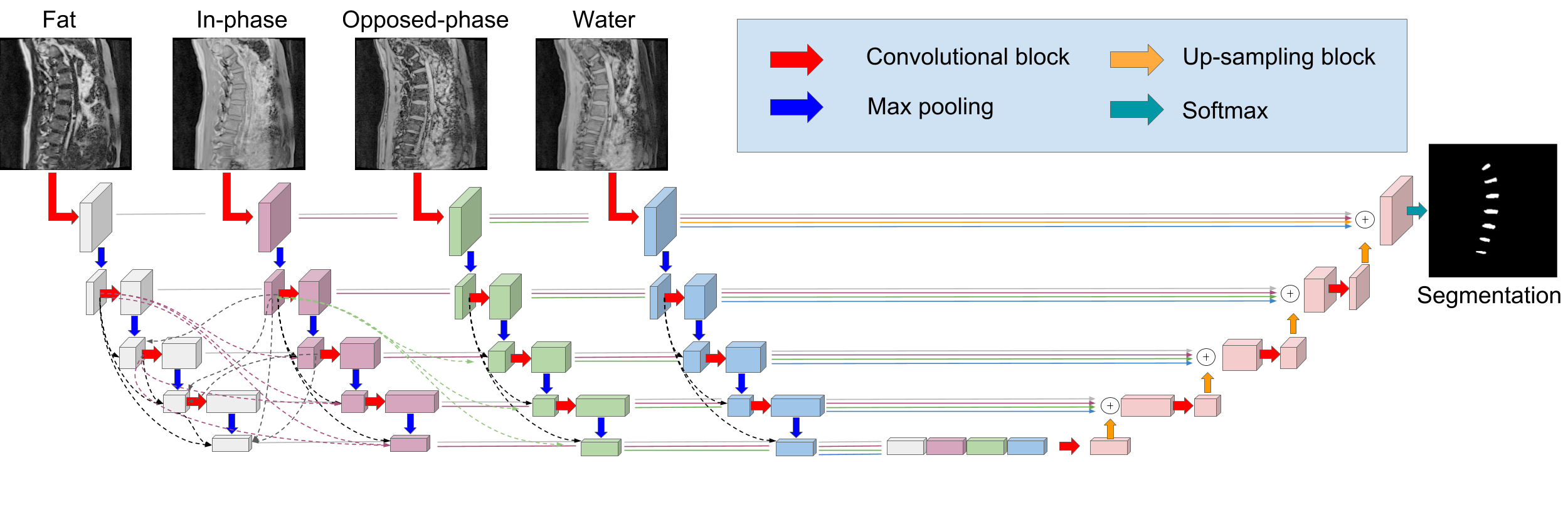}
\caption{Proposed IVD-Net architecture for IVD segmentation in multi-modal images, which extends the traditional UNet. Dotted lines represent some of the dense connectivity patterns adopted in this extended version of UNet.} \label{HDUnet}
\label{fig:mainNet}
\end{figure*}

In order to fully exploit multi-modal data, we adopt the hyper-dense connectivity approach of \cite{dolz2018hyperdense} in the current work. To achieve this dense connectivity pattern, we first create an encoding path composed of multiple streams, each of them processing a different image modality. The main goal of employing separate streams for different modalities is to disentangle information that otherwise would be fused from an early stage, limiting the learning capabilities of the network to capture complex relationships between modalities. The structure of the proposed IVD-Net architecture is depicted in Fig. \ref{fig:mainNet}.

\subsection{Extended Inception module}

Meaningful areas in an image may undergo extremely large variation in size. In our particular case, as 3D segmentation is assessed in a 2D slice-wise manner, the region occupied by the IVD varies from one image to another. For instance, when the 2D sagittal slice corresponds to the center of the vertebral column, every IVD will appear in the image, whereas only one or two IVDs will be present in the image when the sagittal plane is located at extremes. This makes the selection of an accurate and general kernel size difficult. While a smaller kernel is better for local information, a larger kernel can capture information that is distributed globally. This idea is exploited in InceptionNet \cite{szegedy2016rethinking}, where convolutions with multiple kernel sizes operate on the same level. Furthermore, in more recent versions, n$\times$n convolutions are factorized to a combination of 1$\times$n and n$\times$1 convolutions, resulting in a 33$\%$ memory reduction.

\begin{figure}[h!]
\centering
\includegraphics[width=0.55\textwidth]{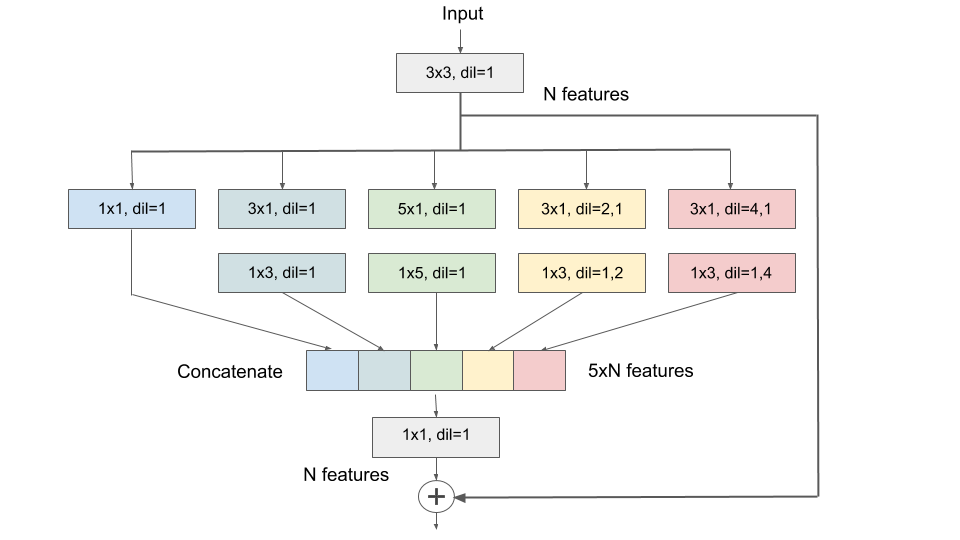}
\caption{Proposed extended inception modules. The module on the left employs standard convolutions while the module on the right adopts the idea of asymmetric convolutions \cite{szegedy2016rethinking}.} 
\label{fig:res-inception-mod}
\end{figure}

To facilitate the learning of multiple contexts, we included two dilated convolutional blocks in parallel to the existing blocks in an inception module. Dilation rates of these blocks are different, which helps learning from different receptive fields, thereby increasing the context of the original inception modules. In addition, we removed max-pooling from the proposed architecture, as dilated convolutions were shown to be a better alternative, which captures more effectively the global context \cite{yu2015multi}. Our extended inception modules are depicted in Fig. \ref{fig:res-inception-mod}.

\subsection{Hyper-Dense connectivity}
\label{ssec:hyper}

Inspired by the recent success of densely connected architectures for medical image segmentation \cite{chen2018mri,dolz2018hyperdense,yu2017automatic}, we adopted hyper-dense connections in the proposed model. The benefits of employing dense connections in the network are four-fold \cite{huang2017densely,dolz2018hyperdense}. First, as demonstrated in \cite{dolz2018hyperdense}, dense connections between multiple streams can better model relationships between different modalities. Second, flow of information and gradients through the entire network is facilitated by the use of direct connections between all layers, which alleviates the problem of vanishing gradient. Third, including short paths to all feature maps in the network introduces an implicit deep supervision. Fourth, dense connections have a regularizing effect, reducing the risk of over-fitting on tasks with smaller training sets.

\subsubsection{Formulation.}

Let $\xx_l$ denote the output of the $l^{th}$ layer, and $H_l$ be a mapping function, which corresponds to a convolution layer followed by a non-linear activation. In standard CNNs, the output of the $l^{th}$ layer is typically obtained from the output of the previous layer $\xx_{l-1}$ as
\begin{equation}
 \xx_l \ = \ H_l\big(\xx_{l-1}\big).
 \label{eq:layer_output}
\end{equation} 
In a densely-connected network, nevertheless, all feature outputs are concatenated in a feed-forward manner, i.e.,
\begin{equation}
 \xx_l \ = \ H_l\big([\xx_{l-1}, \xx_{l-2}, \ldots, \xx_{0}]\big),
 \label{eq:layer_outputDense}
\end{equation}
where $[\ldots]$ denotes a concatenation operation.

In the present work, as in HyperDenseNet \cite{dolz2018isointense,dolz2018hyperdense}, the outputs from previous layers in different streams are also concatenated to form the input of subsequent layers. This connectivity yields a much more powerful feature representation than early or late fusion strategies in a multi-modal context, as the network is capable of learning more complex relationships between the different modalities within and in-between all levels of abstraction. For simplicity, let us consider the scenario with only two modalities. Let $\xx_l^1$ and $\xx_l^2$ denote the outputs of the $l^{th}$ layer in streams 1 and 2, respectively. Then, the output of the $l^{th}$ layer in a given stream $s$ can be defined as
\begin{equation}
 \xx_l^s \ = \ H_l^s\big([\xx_{l-1}^1, \xx_{l-1}^2, \xx_{l-2}^1, \xx_{l-2}^2, \ldots, \xx_{0}^1, \xx_{0}^2]\big).
 \label{eq:layer_HyperDense}
\end{equation}

Furthermore, recent works have found that shuffling and interleaving complete feature maps (or single feature maps elements) in a CNN can improve its performance, as it serves as a strong regularizer \cite{chen2017regularization,zhang2017interleaved,zhang2017shufflenet}. Inspired by this, we concatenate feature maps in a different order for each branch and layer, where the output of the $l^{th}$ layer now becomes
\begin{equation}
  \xx_l^s \ = \ H_l^s\big(\pi_l^s([\xx_{l-1}^1, \xx_{l-1}^2, \xx_{l-2}^1, \xx_{l-2}^2, \ldots, \xx_{0}^1, \xx_{0}^2])\big),
  \label{eq:hyp}
\end{equation}
with $\pi_l^s$ being a function that permutes the feature maps given as input. Thus, in the case of two image modalities, the outputs of the $l^{th}$ layers in both streams can be defined as 
\begin{equation*}
\begin{split}
  \xx_l^1 & \ = \ 
    H_l^1\big([\xx_{l-1}^1, \xx_{l-1}^2, \xx_{l-2}^1, \xx_{l-2}^2, \ldots, \xx_{0}^1, \xx_{0}^2]\big) \\ 
  \xx_l^2 & \ = \ 
    H_l^2\big([\xx_{l-1}^2, \xx_{l-1}^1, \xx_{l-2}^2, \xx_{l-2}^1, \ldots, \xx_{0}^2, \xx_{0}^1])\big.
  \end{split}
\end{equation*}

\begin{figure}[h!]
\centering
\includegraphics[width=.5\textwidth]{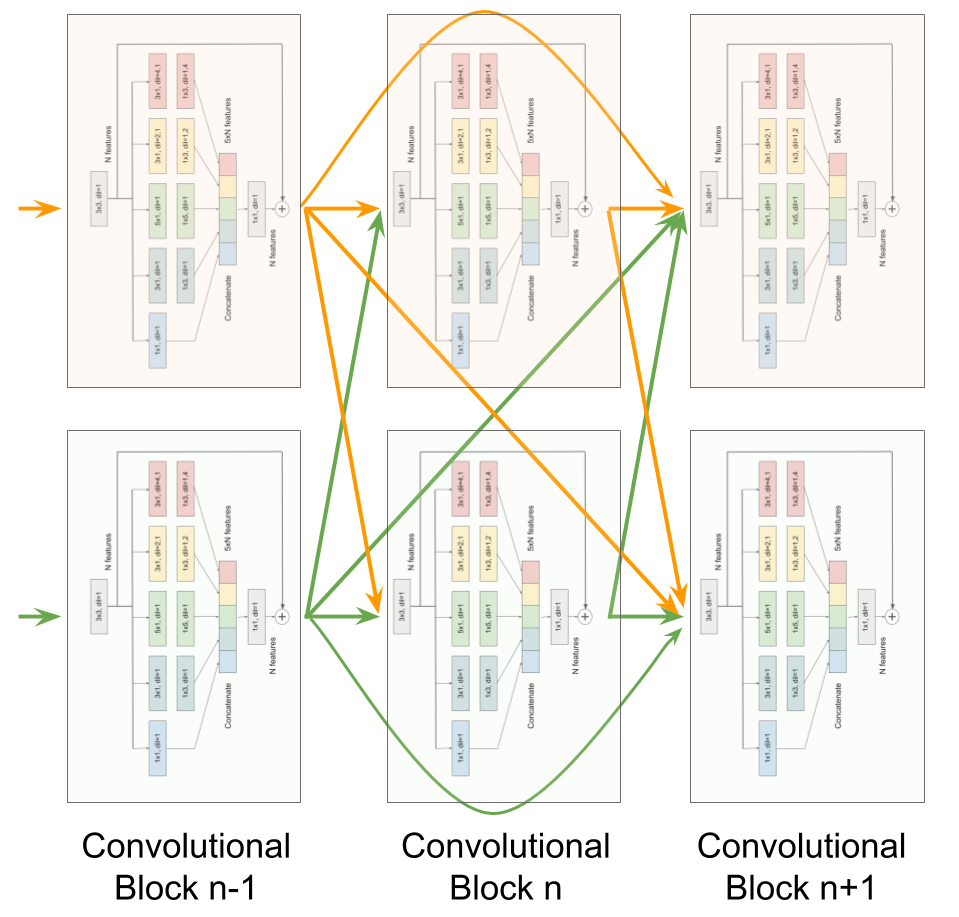}
\caption{Detailed version of a section of the proposed dense connectivity in multi-modal scenarios. For simplicity, two image modalities (in orange and in green) are considered in this example. While boxes represent a complete convolutional block of the proposed type, arrows indicate the connectivity pattern between modules.} 
\label{fig:HD_Detail}
\end{figure}

A detailed example of the adopted hyper-dense connectivity for the case of two image modalities is depicted in Fig. \ref{fig:HD_Detail}. This figure shows a section (only three levels) of a deep CNN where the two image modalities are processed in separated paths and modules are linked in a hyper-dense fashion.

\section{Materials}

\subsection{Dataset}

The provided IVD dataset is composed of 16 3D multi-modal MRI data sets of at least 7 IVDs of the lower spine, collected from 8 subjects in two different stages. Each MRI data set contains four aligned high-resolution 3D volumes: in-phase, opposed-phase, fat and water images. In addition to the MRI images, corresponding reference manual segmentations were provided. More detailed information about the dataset can be found at the IVD website\footnote{https://ivdm3seg.weebly.com}.

\subsection{Evaluation metrics}

Even though segmentation is performed in a 2D-slice fashion, once all the 2D sagittal slices for a given patient have been segmented, they are stacked to reconstruct the original 3D volume. The metrics introduced below are therefore employed to evaluate performance on the whole 3D image. While the first metric is used to evaluate the segmentation accuracy, the second one serves as a measure of localization error.

\paragraph{\textbf{Dice similarity coefficient (DSC).}} We first evaluate performance using Dice similarity coefficient (DSC), which compares volumes based on their overlap. Let $V_\mr{ref}$ and $V_\mr{auto}$ be the reference and automatic segmentations of a given tissue class and for a given subject, respectively. The DSC for this subject is defined as
\begin{equation}
\label{eq:dice}
\mr{DSC}\big(V_\mr{ref}, V_\mr{auto} \big) \ = \ 
 \frac{2 \mid V_\mr{ref} \cap V_\mr{auto}\mid} {\mid V_\mr{ref}\mid +\mid V_\mr{auto}\mid}
\end{equation}

\paragraph{\textbf{Localization distance.}} To evaluate the localization error, we compute the 3D barycenters of ground-truth and predicted IVDs, and measure their Euclidean distance. Results are given in voxels.

\subsection{Implementation details}

\paragraph{\textbf{\textit{Baselines.}}} Several architectures are used to demonstrate the effectiveness of the proposed network. As baselines, we consider two UNet versions, the first one with early fusion and the other with late fusion. In early fusion, following the procedure employed in most works, all MRI image modalities are merged into a single input which is processed through a unique path. In contrast, for late fusion, each MRI modality is processed in a separate stream, and learned features of different modalities are fused in a later stage. In both early and late fusion, the extended inception module of Fig. \ref{fig:res-inception-mod} is employed, however asymmetric convolutions are replaced by standard n$\times$n convolutions in these baselines. Another difference with respect to standard UNet is that feature maps from skip connections are summed before being fed into convolutional modules of the decoding path, instead of being concatenated.

\paragraph{\textbf{\textit{Proposed network.}}} In terms of architecture, the proposed IVD-Net network and the one employed with late fusion strategy are very similar. As introduced in Section \ref{ssec:hyper}, the main difference is that feature maps from previous layers and different paths are concatenated and fed into the subsequent layers, following Eq. (\ref{eq:hyp}). Details of the resulting architecture are provided in Table \ref{ref:arch}. The first version of the proposed network employs the same convolutional module as the two baselines, whereas the second version adopts asymmetric convolutions instead (Fig. \ref{fig:res-inception-mod}).

\begin{table}[ht!]
\centering
\scriptsize
\caption{Layer placement of the proposed hyper-dense connected UNet.}
\renewcommand{\arraystretch}{1.1}
\begin{tabular}{lclccc}
\toprule
\multicolumn{4}{l}{}  & \multicolumn{2}{c}{\textbf{HyperDense connectivity}} \\
\cmidrule(l{2pt}r{2pt}){5-6}
 & & Name &  & \multicolumn{1}{c}{\begin{tabular}[c]{@{}c@{}}Feat maps\\ (input)\end{tabular}} & \multicolumn{1}{c}{\begin{tabular}[c]{@{}c@{}}Feat maps\\ (output)\end{tabular}} \\ 
\midrule
\multirow{7}{*}{\begin{tabular}[c]{@{}l@{}}Encoding\\ Path \\ (each modality)\end{tabular}} 
& & Conv Layer 1 & & 1$\times$256$\times$256 &32$\times$256$\times$256 \\ 
& & Max-pooling 1 & & 32$\times$256$\times$256 & 32$\times$128$\times$128 \\
& & Layer 2  &  &  128$\times$128$\times$128 & 64$\times$128$\times$128  \\
& & Max-pooling 2 & & 64$\times$128$\times$128  & 64$\times$64$\times$64  \\
& & Layer 3 &  & 384$\times$64$\times$64 & 128$\times$64$\times$64   \\
& & Max-pooling 3 &      & 128$\times$64$\times$64 & 128$\times$32$\times$32 \\
& & Layer 4  &  & 896$\times$32$\times$32 &  256$\times$32$\times$32  \\ 
& & Max-pooling 4  &   & 256$\times$32$\times$32 &  256$\times$16$\times$16 \\
\midrule
 & &   Bridge & & 1920$\times$16$\times$16 &  512$\times$16$\times$16 \\
 \midrule
\multirow{8}{*}{\begin{tabular}[c]{@{}l@{}}Decoding\\ Path\end{tabular}} 
& & Up-sample 1  &  & 512$\times$16$\times$16 & 256$\times$32$\times$32 \\
& & Layer 5     &  & 256$\times$32$\times$32  &  256$\times$32$\times$32  \\
& & Up-sample 2  &   & 256$\times$32$\times$32 & 128$\times$64$\times$64   \\
& & Layer 6 & & 128$\times$64$\times$64  &  128$\times$64$\times$64  \\
& & Up-sample 3 & & 128$\times$64$\times$64 &  64$\times$128$\times$128 \\
& & Layer 7 & & 64$\times$128$\times$128 &  64$\times$128$\times$128 \\
& & Up-sample 4 & & 64$\times$128$\times$128 &  32$\times$256$\times$256 \\
& & Layer 8 & & 32$\times$256$\times$256 & 32$\times$256$\times$256   \\
& & Softmax layer & &32$\times$256$\times$256  & 2$\times$256$\times$256  \\ 
\bottomrule
\end{tabular}
\label{ref:arch}
\end{table}

\paragraph{\textbf{\textit{Training.}}} We employed Adam optimizer to train the proposed architectures, with $\beta_1=0.9$ and $\beta_2=0.99$. Training converged after 200 epochs with an initial learning rate of 1$\times$10$^{-4}$, reduced by half after 100 epochs. Four images were used in each mini-batch. The same values for all hyper-parameters were employed across all architectures. Implementation of the analyzed architectures was done in PyTorch and experiments were performed on an NVidia TITAN XP GPU with 16 GBs RAM. While training was done in around 5 hours, inference on a whole 3D volume took in 2-3 seconds on average. Images were normalized between 0 and 1 and no other pre- or post-processing steps were used. Furthermore, no data augmentation was employed to boost the performance of the networks. For all architectures, we used the four MRI modalities provided by the organizers as input. While 13 scans were employed for training 3 scans were used for validation.

\section{Results}\label{sec:experiments}

Quantitative results obtained with the different architectures are reported in Table \ref{table:metrics}. First, we observe that by simply fusing all image modalities at the input of the network provides the lowest mean DSC value. Adopting a late fusion strategy instead of early fusion achieves a mean DSC of 0.9086. Moreover, we see that our hyper-densely connected IVD-Net architecture brings a boost in performance compared to the more `naive' early or late fusion strategies. When employing the extended module with standard convolutions (Fig. \ref{fig:res-inception-mod}), we obtained a mean DSC of 0.9162, whereas the use of asymmetric convolutions in the proposed module provided the best performance in terms of mean DSC. These results are in line with values of localization distance, where the proposed architecture outperforms simpler fusion strategies. Nevertheless, in this case, the proposed network integrating standard convolutions slightly outperforms the architecture with asymmetric convolutions.

\begin{table}[ht!]
\centering
\footnotesize
\caption{Results on validation subjects obtained by the different architectures.}\label{tab1}
\begin{tabular}{lcC{3cm}}
\hline
Architecture &  DSC &  \begin{tabular}[c]{@{}c@{}}Localization
distance\\  (voxels)\end{tabular}\\

\hline
Baseline$\_$EarlyFusion &  0.8981 $\pm$ 0.0293& 0.7701 $\pm$ 1.5872 \\
Baseline$\_$LateFusion &  0.9086 $\pm$ 0.0339& 0.7400 $\pm$ 1.6009\\
IVD-Net &  0.9162 $\pm$ 0.0192& \textbf{0.4145 $\pm$ 0.2698}\\
IVD-Net (asym) &  \textbf{0.9191 $\pm$ 0.0179} & 0.4470 $\pm$ 0.2641\\

\hline
\end{tabular}
\label{table:metrics}
\end{table}



\begin{figure}[h!]
\centering
\includegraphics[width=0.5\textwidth]{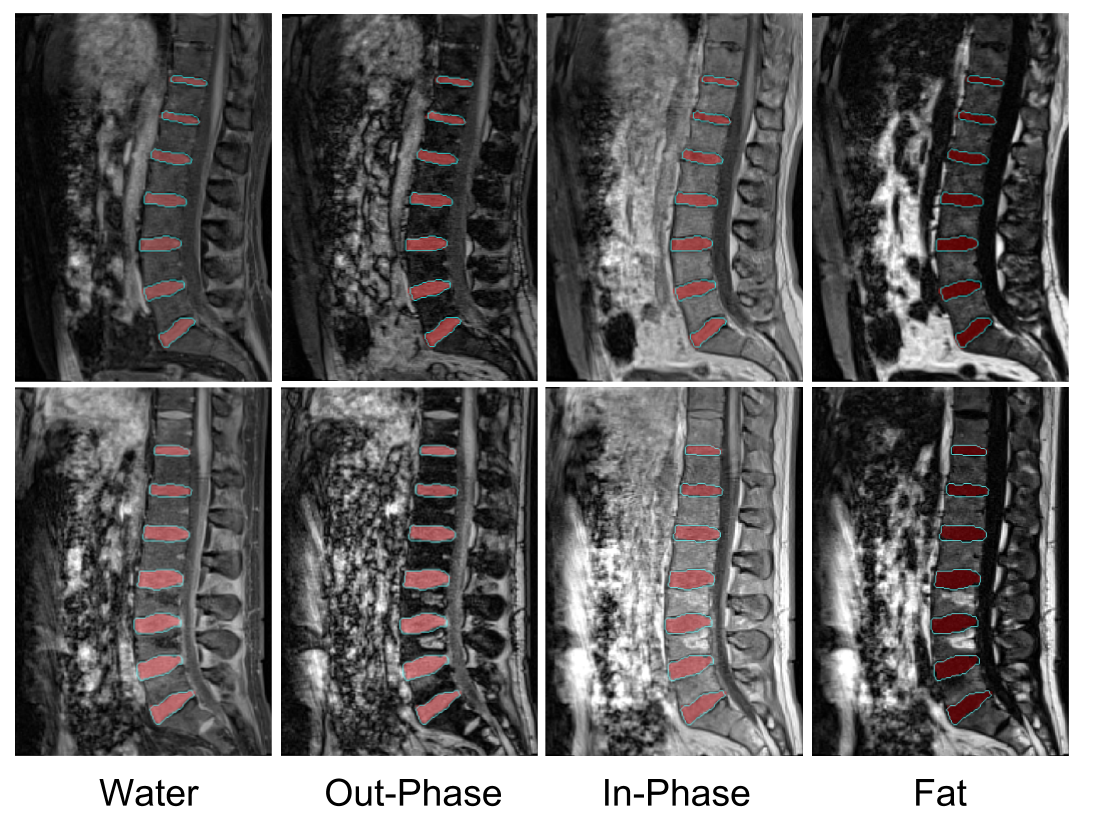}
\caption{Visual results for two subjects of the validation set. While the area in red represents the ground truth, bluish contours depict the automatic contours by our IVD-Net (asym) method in the different image modalities.} 
\label{fig:ISLES_Res}
\end{figure}

\begin{figure}[h!]
\centering
\includegraphics[width=0.5\textwidth]{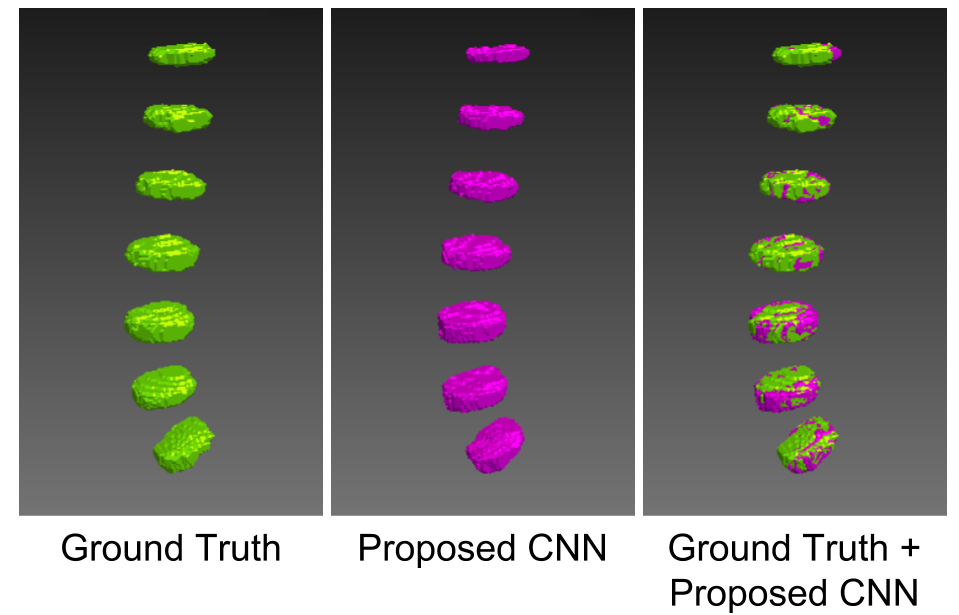}
\caption{3D visualization of the ground truth, segmentation achieved by the proposed network and the combination of both for a subject on the validation set.} 
\label{fig:3D}
\end{figure}

Qualitative evaluation of the proposed IVD-Net architecture is assessed in Fig. \ref{fig:ISLES_Res} and \ref{fig:3D}. First, ground truth and automatic contours obtained with IVD-Net are depicted on the sagittal plane in Fig. \ref{fig:ISLES_Res} for two validation subjects. Then, 3D rendered volumes for the ground truth and CNN segmentation are compared in Fig. \ref{fig:3D}. In both figures, we can see that the segmentation obtained by our architecture is very close to the manual annotated data, which aligns with the quantitative results in Table \ref{table:metrics}.


\section{Discussion}

\begin{figure*}[ht!]
\centering
\includegraphics[width=1\textwidth]{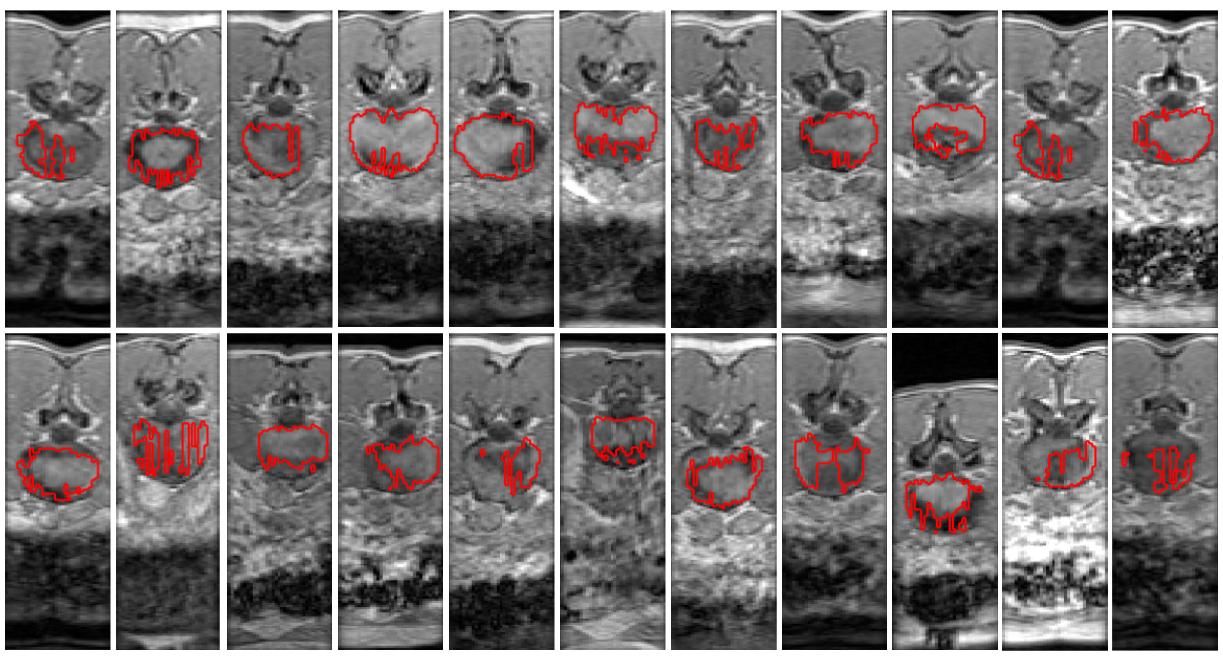}
\caption{Examples of manual annotations from the training set seen on axial slices.}
\label{fig:badManual}
\end{figure*}

We have presented an architecture called IVD-Net that can efficiently leverage information from multiple image modalities for inter-vertebral disc segmentation. Following recent research on multi-modal image segmentation \cite{dolz2018dense,dolz2018hyperdense}, our architecture adopts dense connectivity between multiple paths in the encoding section, each of them processing single modalities. Specifically, convolutional layers in any stream receive as input the features maps of all previous layers in the same stream as well as from other streams. 


We have demonstrated that naive feature fusion strategies, such as simply merging information at an early or late stage, may be insufficient to fully exploit information in multi-modal scenarios. By allowing the network to learn how to combine learned features from separate modalities, it can capture more complex relationships between multiple sources. This improves its representation power, which ultimately results in a boost on performance. These findings are in line with recent works on multi-modal image segmentation \cite{dolz2018isointense,dolz2018hyperdense,nie2016fully}. For example, high-level features were combined at a late stage in \cite{nie2016fully}, outperforming an early fusion strategy in the context of infant brain segmentation. In a recent work, we demonstrated that adopting more complex fusion techniques, referred to as hyper-dense connectivity, surpasses the performance of other features fusion strategies in the challenging tasks of infant and adult brain tissue segmentation \cite{dolz2018isointense,dolz2018hyperdense}.

Even though considering 3D context typically helps improve performance, we treated each volume as a stack of 2D sagittal slices (see Fig. \ref{fig:badManual}). The main reason for this is that manual segmentations provided in this challenge were performed slice-wise in the sagittal plane. Thus, when looking at these annotations in the axial plane, a sharp contour is observed. As CNNs will generally provide a smooth contour, we assumed that tackling this problem as a 3D task would have led to lower values during evaluation. Furthermore, IVD localization is assessed after volumetric segmentation is done. This means that the process of localization itself is not optimized during training. A possible solution to overcome this limitation in the future might be to investigate multi-task architectures that can be trained end-to-end, so that both localization and segmentation tasks can be jointly optimized.

\ifCLASSOPTIONcompsoc
 \section*{Acknowledgments}
\else
 \section*{Acknowledgment}
\fi

This work is supported by the National Science and Engineering Research Council of Canada (NSERC), discovery grant program, and by the ETS Research Chair on Artificial Intelligence in Medical Imaging.

\ifCLASSOPTIONcaptionsoff
 \newpage
\fi

\bibliographystyle{IEEEtran}
\bibliography{biblio}

\begin{thebibliography}{10}
\providecommand{\url}[1]{#1}
\csname url@samestyle\endcsname
\providecommand{\newblock}{\relax}
\providecommand{\bibinfo}[2]{#2}
\providecommand{\BIBentrySTDinterwordspacing}{\spaceskip=0pt\relax}
\providecommand{\BIBentryALTinterwordstretchfactor}{4}
\providecommand{\BIBentryALTinterwordspacing}{\spaceskip=\fontdimen2\font plus
\BIBentryALTinterwordstretchfactor\fontdimen3\font minus
  \fontdimen4\font\relax}
\providecommand{\BIBforeignlanguage}[2]{{%
\expandafter\ifx\csname l@#1\endcsname\relax
\typeout{** WARNING: IEEEtran.bst: No hyphenation pattern has been}%
\typeout{** loaded for the language `#1'. Using the pattern for}%
\typeout{** the default language instead.}%
\else
\language=\csname l@#1\endcsname
\fi
#2}}
\providecommand{\BIBdecl}{\relax}
\BIBdecl

\bibitem{dolz2018hyperdense}
J.~Dolz, K.~Gopinath, J.~Yuan, H.~Lombaert, C.~Desrosiers, and I.~Ben~Ayed,
  ``Hyper{D}ense-{N}et: {A} hyper-densely connected {CNN} for multi-modal image
  segmentation,'' \emph{arXiv preprint arXiv:1804.02967}, 2018.

\bibitem{szegedy2016rethinking}
C.~Szegedy, V.~Vanhoucke, S.~Ioffe, J.~Shlens, and Z.~Wojna, ``Rethinking the
  inception architecture for computer vision,'' in \emph{CVPR}, 2016, pp.
  2818--2826.

\bibitem{an2004introduction}
H.~S. An, P.~A. Anderson, V.~M. Haughton, J.~C. Iatridis, J.~D. Kang, J.~C.
  Lotz, R.~N. Natarajan, T.~R. Oegema~Jr, P.~Roughley, L.~A. Setton
  \emph{et~al.}, ``Introduction: disc degeneration: summary,'' \emph{Spine},
  vol.~29, no.~23, pp. 2677--2678, 2004.

\bibitem{wieser2011cost}
S.~Wieser, B.~Horisberger, S.~Schmidhauser, C.~Eisenring, U.~Br{\"u}gger,
  A.~Ruckstuhl, J.~Dietrich, A.~F. Mannion, A.~Elfering, {\"O}.~Tamcan
  \emph{et~al.}, ``Cost of low back pain in switzerland in 2005,'' \emph{The
  European Journal of Health Economics}, vol.~12, no.~5, pp. 455--467, 2011.

\bibitem{hamanishi1994cross}
C.~Hamanishi, N.~Matukura, M.~Fujita, M.~Tomihara, and S.~Tanaka,
  ``Cross-sectional area of the stenotic lumbar dural tube measured from the
  transverse views of magnetic resonance imaging.'' \emph{Journal of spinal
  disorders}, vol.~7, no.~5, pp. 388--393, 1994.

\bibitem{ayed2011graph}
I.~B. Ayed, K.~Punithakumar, G.~Garvin, W.~Romano, and S.~Li, ``Graph cuts with
  invariant object-interaction priors: application to intervertebral disc
  segmentation,'' in \emph{Biennial International Conference on Information
  Processing in Medical Imaging}.\hskip 1em plus 0.5em minus 0.4em\relax
  Springer, 2011, pp. 221--232.

\bibitem{chen2015localization}
C.~Chen, D.~Belavy, W.~Yu, C.~Chu, G.~Armbrecht, M.~Bansmann, D.~Felsenberg,
  and G.~Zheng, ``Localization and segmentation of {3D} intervertebral discs in
  {MR} images by data driven estimation,'' \emph{IEEE transactions on medical
  imaging}, vol.~34, no.~8, pp. 1719--1729, 2015.

\bibitem{chen20163d}
H.~Chen, Q.~Dou, X.~Wang, J.~Qin, J.~C. Cheng, and P.-A. Heng, ``3{D} fully
  convolutional networks for intervertebral disc localization and
  segmentation,'' in \emph{International Conference on Medical Imaging and
  Virtual Reality}.\hskip 1em plus 0.5em minus 0.4em\relax Springer, 2016, pp.
  375--382.

\bibitem{ji2016automated}
X.~Ji, G.~Zheng, D.~Belavy, and D.~Ni, ``Automated intervertebral disc
  segmentation using deep convolutional neural networks,'' in
  \emph{International Workshop on Computational Methods and Clinical
  Applications for Spine Imaging}.\hskip 1em plus 0.5em minus 0.4em\relax
  Springer, 2016, pp. 38--48.

\bibitem{kim2018fine}
S.~Kim, W.~Bae, K.~Masuda, C.~Chung, and D.~Hwang, ``Fine-grain segmentation of
  the intervertebral discs from {MR} spine images using deep convolutional
  neural networks: {BSU}-{N}et,'' \emph{Applied Sciences}, vol.~8, no.~9, p.
  1656, 2018.

\bibitem{zeng2017dsms}
G.~Zeng and G.~Zheng, ``{DSMS-FCN}: A deeply supervised multi-scale fully
  convolutional network for automatic segmentation of intervertebral disc in
  3{D} {MR} images,'' in \emph{International Workshop and Challenge on
  Computational Methods and Clinical Applications in Musculoskeletal
  Imaging}.\hskip 1em plus 0.5em minus 0.4em\relax Springer, 2017, pp.
  148--159.

\bibitem{zheng2017evaluation}
G.~Zheng, C.~Chu, D.~L. Belav{\`y}, B.~Ibragimov, R.~Korez, T.~Vrtovec,
  H.~Hutt, R.~Everson, J.~Meakin, I.~L. Andrade \emph{et~al.}, ``Evaluation and
  comparison of 3{D} intervertebral disc localization and segmentation methods
  for {3D} {T}2 {MR} data: {A} grand challenge,'' \emph{Medical image
  analysis}, vol.~35, pp. 327--344, 2017.

\bibitem{zhang2015deep}
W.~Zhang, R.~Li, H.~Deng, L.~Wang, W.~Lin, S.~Ji, and D.~Shen, ``Deep
  convolutional neural networks for multi-modality isointense infant brain
  image segmentation,'' \emph{NeuroImage}, vol. 108, pp. 214--224, 2015.

\bibitem{moeskops2016automatic}
P.~Moeskops, M.~A. Viergever, A.~M. Mendrik, L.~S. de~Vries, M.~J. Benders, and
  I.~I{\v{s}}gum, ``Automatic segmentation of {MR} brain images with a
  convolutional neural network,'' \emph{IEEE Transactions on Medical Imaging},
  vol.~35, no.~5, pp. 1252--1261, 2016.

\bibitem{kamnitsas2017efficient}
K.~Kamnitsas, C.~Ledig, V.~F. Newcombe, J.~P. Simpson, A.~D. Kane, D.~K. Menon,
  D.~Rueckert, and B.~Glocker, ``Efficient multi-scale {3D} {CNN} with fully
  connected {CRF} for accurate brain lesion segmentation,'' \emph{Medical image
  analysis}, vol.~36, pp. 61--78, 2017.

\bibitem{dolz2017deep}
J.~Dolz, C.~Desrosiers, L.~Wang, J.~Yuan, D.~Shen, and I.~Ben~Ayed, ``Deep
  {CNN} ensembles and suggestive annotations for infant brain {MRI}
  segmentation,'' \emph{arXiv preprint arXiv:1712.05319}, 2017.

\bibitem{valverde2017improving}
S.~Valverde, M.~Cabezas, E.~Roura, S.~Gonz{\'a}lez-Vill{\`a}, D.~Pareto, J.~C.
  Vilanova, L.~Rami{\'o}-Torrent{\`a}, {\`A}.~Rovira, A.~Oliver, and
  X.~Llad{\'o}, ``Improving automated multiple sclerosis lesion segmentation
  with a cascaded {3D} convolutional neural network approach,''
  \emph{NeuroImage}, vol. 155, pp. 159--168, 2017.

\bibitem{Srivastava14}
N.~Srivastava and R.~Salakhutdinov, ``Multimodal learning with deep boltzmann
  machines,'' \emph{Journal of Machine Learning Research}, vol.~15, pp.
  2949--2980, 2014.

\bibitem{nie2016fully}
D.~Nie, L.~Wang, Y.~Gao, and D.~Sken, ``Fully convolutional networks for
  multi-modality isointense infant brain image segmentation,'' in \emph{13th
  International Symposium on Biomedical Imaging (ISBI), 2016}.\hskip 1em plus
  0.5em minus 0.4em\relax IEEE, 2016, pp. 1342--1345.

\bibitem{aygun2018multi}
M.~Ayg{\"u}n, Y.~H. {\c{S}}ahin, and G.~{\"U}nal, ``Multi modal convolutional
  neural networks forbrain tumor segmentation,'' \emph{arXiv preprint
  arXiv:1809.06191}, 2018.

\bibitem{dolz2018isointense}
J.~Dolz, I.~Ben~Ayed, J.~Yuan, and C.~Desrosiers, ``Isointense infant brain
  segmentation with a hyper-dense connected convolutional neural network,'' in
  \emph{Biomedical Imaging (ISBI 2018), 2018 IEEE 15th International Symposium
  on}.\hskip 1em plus 0.5em minus 0.4em\relax IEEE, 2018, pp. 616--620.

\bibitem{li20183d}
X.~Li, Q.~Dou, H.~Chen, C.-W. Fu, X.~Qi, D.~L. Belav{\`y}, G.~Armbrecht,
  D.~Felsenberg, G.~Zheng, and P.-A. Heng, ``3{D} multi-scale {FCN} with random
  modality voxel dropout learning for intervertebral disc localization and
  segmentation from multi-modality {MR} images,'' \emph{Medical image
  analysis}, vol.~45, pp. 41--54, 2018.

\bibitem{dolz2018dense}
J.~Dolz, I.~Ben~Ayed, and C.~Desrosiers, ``Dense multi-path {U}-{N}et for
  ischemic stroke lesion segmentation in multiple image modalities,''
  \emph{arXiv preprint arXiv:1810.07003}, 2018.

\bibitem{ronneberger2015u}
O.~Ronneberger, P.~Fischer, and T.~Brox, ``U-net: Convolutional networks for
  biomedical image segmentation,'' in \emph{MICCAI}.\hskip 1em plus 0.5em minus
  0.4em\relax Springer, 2015, pp. 234--241.

\bibitem{yu2015multi}
F.~Yu and V.~Koltun, ``Multi-scale context aggregation by dilated
  convolutions,'' \emph{arXiv preprint arXiv:1511.07122}, 2015.

\bibitem{chen2018mri}
L.~Chen, Y.~Wu, A.~M. DSouza, A.~Z. Abidin, A.~Wism{\"u}ller, and C.~Xu,
  ``{MRI} tumor segmentation with densely connected {3D} {CNN},'' in
  \emph{Medical Imaging 2018: Image Processing}.\hskip 1em plus 0.5em minus
  0.4em\relax International Society for Optics and Photonics, 2018.

\bibitem{yu2017automatic}
L.~Yu, J.-Z. Cheng, Q.~Dou, X.~Yang, H.~Chen, J.~Qin, and P.-A. Heng,
  ``Automatic {3D} cardiovascular {MR} segmentation with densely-connected
  volumetric convnets,'' in \emph{MICCAI}.\hskip 1em plus 0.5em minus
  0.4em\relax Springer, 2017, pp. 287--295.

\bibitem{huang2017densely}
G.~Huang, Z.~Liu, L.~Van Der~Maaten, and K.~Q. Weinberger, ``Densely connected
  convolutional networks.'' in \emph{CVPR}, vol.~1, no.~2, 2017, p.~3.

\bibitem{chen2017regularization}
Y.~Chen, H.~Wang, and Y.~Long, ``Regularization of convolutional neural
  networks using shufflenode,'' in \emph{Multimedia and Expo (ICME), 2017 IEEE
  International Conference on}.\hskip 1em plus 0.5em minus 0.4em\relax IEEE,
  2017, pp. 355--360.

\bibitem{zhang2017interleaved}
T.~Zhang, G.-J. Qi, B.~Xiao, and J.~Wang, ``Interleaved group convolutions,''
  in \emph{CVPR}, 2017, pp. 4373--4382.

\bibitem{zhang2017shufflenet}
X.~Zhang, X.~Zhou, M.~Lin, and J.~Sun, ``Shufflenet: An extremely efficient
  convolutional neural network for mobile devices,'' \emph{arXiv preprint
  arXiv:1707.01083}, 2017.

\end{thebibliography}


\begin{thebibliography}{71}%
\makeatletter
\providecommand \@ifxundefined [1]{%
 \@ifx{#1\undefined}
}%
\providecommand \@ifnum [1]{%
 \ifnum #1\expandafter \@firstoftwo
 \else \expandafter \@secondoftwo
 \fi
}%
\providecommand \@ifx [1]{%
 \ifx #1\expandafter \@firstoftwo
 \else \expandafter \@secondoftwo
 \fi
}%
\providecommand \natexlab [1]{#1}%
\providecommand \enquote  [1]{``#1''}%
\providecommand \bibnamefont  [1]{#1}%
\providecommand \bibfnamefont [1]{#1}%
\providecommand \citenamefont [1]{#1}%
\providecommand \href@noop [0]{\@secondoftwo}%
\providecommand \href [0]{\begingroup \@sanitize@url \@href}%
\providecommand \@href[1]{\@@startlink{#1}\@@href}%
\providecommand \@@href[1]{\endgroup#1\@@endlink}%
\providecommand \@sanitize@url [0]{\catcode `\\12\catcode `\$12\catcode
  `\&12\catcode `\#12\catcode `\^12\catcode `\_12\catcode `\%12\relax}%
\providecommand \@@startlink[1]{}%
\providecommand \@@endlink[0]{}%
\providecommand \url  [0]{\begingroup\@sanitize@url \@url }%
\providecommand \@url [1]{\endgroup\@href {#1}{\urlprefix }}%
\providecommand \urlprefix  [0]{URL }%
\providecommand \Eprint [0]{\href }%
\providecommand \doibase [0]{http://dx.doi.org/}%
\providecommand \selectlanguage [0]{\@gobble}%
\providecommand \bibinfo  [0]{\@secondoftwo}%
\providecommand \bibfield  [0]{\@secondoftwo}%
\providecommand \translation [1]{[#1]}%
\providecommand \BibitemOpen [0]{}%
\providecommand \bibitemStop [0]{}%
\providecommand \bibitemNoStop [0]{.\EOS\space}%
\providecommand \EOS [0]{\spacefactor3000\relax}%
\providecommand \BibitemShut  [1]{\csname bibitem#1\endcsname}%
\let\auto@bib@innerbib\@empty
\bibitem [{\citenamefont {Antoni}\ \emph {et~al.}(2017)\citenamefont {Antoni},
  \citenamefont {Ferlay}, \citenamefont {Soerjomataram}, \citenamefont {Znaor},
  \citenamefont {Jemal},\ and\ \citenamefont {Bray}}]{Antoni2017Bladder}%
  \BibitemOpen
  \bibfield  {author} {\bibinfo {author} {\bibfnamefont {S}~\bibnamefont
  {Antoni}}, \bibinfo {author} {\bibfnamefont {J}~\bibnamefont {Ferlay}},
  \bibinfo {author} {\bibfnamefont {I}~\bibnamefont {Soerjomataram}}, \bibinfo
  {author} {\bibfnamefont {A}~\bibnamefont {Znaor}}, \bibinfo {author}
  {\bibfnamefont {A}~\bibnamefont {Jemal}}, \ and\ \bibinfo {author}
  {\bibfnamefont {F}~\bibnamefont {Bray}},\ }\bibfield  {title} {\enquote
  {\bibinfo {title} {Bladder cancer incidence and mortality: {A} global
  overview and recent trends},}\ }\href@noop {} {\bibfield  {journal} {\bibinfo
   {journal} {European Urology}\ }\textbf {\bibinfo {volume} {71}},\ \bibinfo
  {pages} {96} (\bibinfo {year} {2017})}\BibitemShut {NoStop}%
\bibitem [{\citenamefont {Woo}\ \emph {et~al.}(2017)\citenamefont {Woo},
  \citenamefont {Suh}, \citenamefont {Kim}, \citenamefont {Cho},\ and\
  \citenamefont {Kim}}]{Woo2017Diagnostic}%
  \BibitemOpen
  \bibfield  {author} {\bibinfo {author} {\bibfnamefont {S.}~\bibnamefont
  {Woo}}, \bibinfo {author} {\bibfnamefont {C.~H.}\ \bibnamefont {Suh}},
  \bibinfo {author} {\bibfnamefont {S.~Y.}\ \bibnamefont {Kim}}, \bibinfo
  {author} {\bibfnamefont {J.~Y.}\ \bibnamefont {Cho}}, \ and\ \bibinfo
  {author} {\bibfnamefont {S.~H.}\ \bibnamefont {Kim}},\ }\bibfield  {title}
  {\enquote {\bibinfo {title} {Diagnostic performance of {MRI} for prediction
  of muscle-invasiveness of bladder cancer: {A} systematic review and
  meta-analysis},}\ }\href@noop {} {\bibfield  {journal} {\bibinfo  {journal}
  {European Journal of Radiology}\ ,\ \bibinfo {pages} {46--55}} (\bibinfo
  {year} {2017})}\BibitemShut {NoStop}%
\bibitem [{\citenamefont {Alfred}\ \emph {et~al.}(2016)\citenamefont {Alfred},
  \citenamefont {Lebret}, \citenamefont {Compérat}, \citenamefont {Cowan},
  \citenamefont {De}, \citenamefont {Bruins}, \citenamefont {Hernández},
  \citenamefont {Espinós}, \citenamefont {Dunn},\ and\ \citenamefont
  {Rouanne}}]{Alfred2016Updated}%
  \BibitemOpen
  \bibfield  {author} {\bibinfo {author} {\bibfnamefont {Witjes~J}\
  \bibnamefont {Alfred}}, \bibinfo {author} {\bibfnamefont {T}~\bibnamefont
  {Lebret}}, \bibinfo {author} {\bibfnamefont {E.~M.}\ \bibnamefont
  {Compérat}}, \bibinfo {author} {\bibfnamefont {N.~C.}\ \bibnamefont
  {Cowan}}, \bibinfo {author} {\bibfnamefont {Santis~M}\ \bibnamefont {De}},
  \bibinfo {author} {\bibfnamefont {H.~M.}\ \bibnamefont {Bruins}}, \bibinfo
  {author} {\bibfnamefont {V}~\bibnamefont {Hernández}}, \bibinfo {author}
  {\bibfnamefont {E.~L.}\ \bibnamefont {Espinós}}, \bibinfo {author}
  {\bibfnamefont {J}~\bibnamefont {Dunn}}, \ and\ \bibinfo {author}
  {\bibfnamefont {M}~\bibnamefont {Rouanne}},\ }\bibfield  {title} {\enquote
  {\bibinfo {title} {Updated 2016 {EAU} guidelines on muscle-invasive and
  metastatic bladder cancer},}\ }\href@noop {} {\bibfield  {journal} {\bibinfo
  {journal} {European Urology}\ } (\bibinfo {year} {2016})}\BibitemShut
  {NoStop}%
\bibitem [{\citenamefont {{American Cancer
  Society}}(2016)}]{AmericanCancerSociety2016}%
  \BibitemOpen
  \bibfield  {author} {\bibinfo {author} {\bibnamefont {{American Cancer
  Society}}},\ }\bibfield  {title} {\enquote {\bibinfo {title} {{Cancer Facts
  {\&} Figures 2016}},}\ }\href {\doibase 10.1097/01.NNR.0000289503.22414.79}
  {\bibfield  {journal} {\bibinfo  {journal} {Cancer Facts {\&} Figures 2016}\
  ,\ \bibinfo {pages} {1--9}} (\bibinfo {year} {2016})},\ \Eprint
  {http://arxiv.org/abs/NIHMS150003} {arXiv:NIHMS150003} \BibitemShut {NoStop}%
\bibitem [{\citenamefont {Kamat}\ \emph {et~al.}(2016)\citenamefont {Kamat},
  \citenamefont {Hahn}, \citenamefont {Efstathiou}, \citenamefont {Lerner},
  \citenamefont {Malmstr{\"{o}}m}, \citenamefont {Choi}, \citenamefont {Guo},\
  and\ \citenamefont {Lotan}}]{Kamat2016}%
  \BibitemOpen
  \bibfield  {author} {\bibinfo {author} {\bibfnamefont {Ashish~M}\
  \bibnamefont {Kamat}}, \bibinfo {author} {\bibfnamefont {Noah~M}\
  \bibnamefont {Hahn}}, \bibinfo {author} {\bibfnamefont {Jason~A}\
  \bibnamefont {Efstathiou}}, \bibinfo {author} {\bibfnamefont {Seth~P}\
  \bibnamefont {Lerner}}, \bibinfo {author} {\bibfnamefont {Per-uno}\
  \bibnamefont {Malmstr{\"{o}}m}}, \bibinfo {author} {\bibfnamefont
  {Woonyoung}\ \bibnamefont {Choi}}, \bibinfo {author} {\bibfnamefont
  {Charles~C}\ \bibnamefont {Guo}}, \ and\ \bibinfo {author} {\bibfnamefont
  {Yair}\ \bibnamefont {Lotan}},\ }\bibfield  {title} {\enquote {\bibinfo
  {title} {{Bladder cancer}},}\ }\href {\doibase 10.1016/S0140-6736(16)30512-8}
  {\bibfield  {journal} {\bibinfo  {journal} {The Lancet}\ }\textbf {\bibinfo
  {volume} {388}} (\bibinfo {year} {2016}),\
  10.1016/S0140-6736(16)30512-8}\BibitemShut {NoStop}%
\bibitem [{\citenamefont {Choi}\ \emph {et~al.}(2015)\citenamefont {Choi},
  \citenamefont {Porten}, \citenamefont {Kim}, \citenamefont {Willis},
  \citenamefont {Plimack}, \citenamefont {Roth}, \citenamefont {Cheng},
  \citenamefont {Tran}, \citenamefont {Lee}, \citenamefont {Melquist},
  \citenamefont {Bondaruk}, \citenamefont {Majewski}, \citenamefont {Zhang},
  \citenamefont {Pretzsch},\ and\ \citenamefont {Baggerly}}]{Choi2015}%
  \BibitemOpen
  \bibfield  {author} {\bibinfo {author} {\bibfnamefont {Woonyoung}\
  \bibnamefont {Choi}}, \bibinfo {author} {\bibfnamefont {Sima}\ \bibnamefont
  {Porten}}, \bibinfo {author} {\bibfnamefont {Seungchan}\ \bibnamefont {Kim}},
  \bibinfo {author} {\bibfnamefont {Daniel}\ \bibnamefont {Willis}}, \bibinfo
  {author} {\bibfnamefont {Elizabeth~R}\ \bibnamefont {Plimack}}, \bibinfo
  {author} {\bibfnamefont {Beat}\ \bibnamefont {Roth}}, \bibinfo {author}
  {\bibfnamefont {Tiewei}\ \bibnamefont {Cheng}}, \bibinfo {author}
  {\bibfnamefont {Mai}\ \bibnamefont {Tran}}, \bibinfo {author} {\bibfnamefont
  {I-ling}\ \bibnamefont {Lee}}, \bibinfo {author} {\bibfnamefont {Jonathan}\
  \bibnamefont {Melquist}}, \bibinfo {author} {\bibfnamefont {Jolanta}\
  \bibnamefont {Bondaruk}}, \bibinfo {author} {\bibfnamefont {Tadeusz}\
  \bibnamefont {Majewski}}, \bibinfo {author} {\bibfnamefont {Shizhen}\
  \bibnamefont {Zhang}}, \bibinfo {author} {\bibfnamefont {Shanna}\
  \bibnamefont {Pretzsch}}, \ and\ \bibinfo {author} {\bibfnamefont {Keith}\
  \bibnamefont {Baggerly}},\ }\bibfield  {title} {\enquote {\bibinfo {title}
  {{Identification of distinct basal and luminal subtypes of muscle-invasive
  bladder cancer with different sensitivities to frontline chemotherapy}},}\
  }\href {\doibase 10.1016/j.ccr.2014.01.009.Identification} {\bibfield
  {journal} {\bibinfo  {journal} {Cancer Cell}\ }\textbf {\bibinfo {volume}
  {25}},\ \bibinfo {pages} {152--165} (\bibinfo {year} {2015})}\BibitemShut
  {NoStop}%
\bibitem [{\citenamefont {Knowles}\ and\ \citenamefont
  {Hurst}(2015)}]{Knowles2015}%
  \BibitemOpen
  \bibfield  {author} {\bibinfo {author} {\bibfnamefont {Margaret~A}\
  \bibnamefont {Knowles}}\ and\ \bibinfo {author} {\bibfnamefont {Carolyn~D}\
  \bibnamefont {Hurst}},\ }\bibfield  {title} {\enquote {\bibinfo {title}
  {{Molecular biology of bladder cancer : new insights into pathogenesis and
  clinical diversity}},}\ }\href {\doibase 10.1038/nrc3817} {\bibfield
  {journal} {\bibinfo  {journal} {Nature Publishing Group}\ }\textbf {\bibinfo
  {volume} {15}},\ \bibinfo {pages} {25--41} (\bibinfo {year}
  {2015})}\BibitemShut {NoStop}%
\bibitem [{\citenamefont {{Cancer Genome Atlas Research Network and
  others}}(2014)}]{Cancer2014}%
  \BibitemOpen
  \bibfield  {author} {\bibinfo {author} {\bibnamefont {{Cancer Genome Atlas
  Research Network and others}}},\ }\bibfield  {title} {\enquote {\bibinfo
  {title} {{Comprehensive molecular characterization of urothelial bladder
  carcinoma}},}\ }\href {\doibase 10.1038/nature12965} {\bibfield  {journal}
  {\bibinfo  {journal} {Nature}\ }\textbf {\bibinfo {volume} {507}},\ \bibinfo
  {pages} {315--322} (\bibinfo {year} {2014})}\BibitemShut {NoStop}%
\bibitem [{\citenamefont {Duan}\ \emph {et~al.}(2010)\citenamefont {Duan},
  \citenamefont {Liang}, \citenamefont {Bao}, \citenamefont {Zhu},
  \citenamefont {Wang}, \citenamefont {Zhang}, \citenamefont {Chen},\ and\
  \citenamefont {Lu}}]{Duan2010}%
  \BibitemOpen
  \bibfield  {author} {\bibinfo {author} {\bibfnamefont {Chaijie}\ \bibnamefont
  {Duan}}, \bibinfo {author} {\bibfnamefont {Zhengrong}\ \bibnamefont {Liang}},
  \bibinfo {author} {\bibfnamefont {Shangliang}\ \bibnamefont {Bao}}, \bibinfo
  {author} {\bibfnamefont {Hongbin}\ \bibnamefont {Zhu}}, \bibinfo {author}
  {\bibfnamefont {Su}~\bibnamefont {Wang}}, \bibinfo {author} {\bibfnamefont
  {Guangxiang}\ \bibnamefont {Zhang}}, \bibinfo {author} {\bibfnamefont
  {John~J}\ \bibnamefont {Chen}}, \ and\ \bibinfo {author} {\bibfnamefont
  {Hongbing}\ \bibnamefont {Lu}},\ }\bibfield  {title} {\enquote {\bibinfo
  {title} {{A coupled level set framework for bladder wall segmentation with
  application to MR cystography}},}\ }\href {\doibase 10.1109/TMI.2009.2039756}
  {\bibfield  {journal} {\bibinfo  {journal} {IEEE Transactions on Medical
  Imaging}\ }\textbf {\bibinfo {volume} {29}},\ \bibinfo {pages} {903--915}
  (\bibinfo {year} {2010})}\BibitemShut {NoStop}%
\bibitem [{\citenamefont {Qin}\ \emph {et~al.}(2014{\natexlab{a}})\citenamefont
  {Qin}, \citenamefont {Li}, \citenamefont {Liu}, \citenamefont {Lu},\ and\
  \citenamefont {Yan}}]{qin2014adaptive}%
  \BibitemOpen
  \bibfield  {author} {\bibinfo {author} {\bibfnamefont {Xianjing}\
  \bibnamefont {Qin}}, \bibinfo {author} {\bibfnamefont {Xuelong}\ \bibnamefont
  {Li}}, \bibinfo {author} {\bibfnamefont {Yang}\ \bibnamefont {Liu}}, \bibinfo
  {author} {\bibfnamefont {Hongbing}\ \bibnamefont {Lu}}, \ and\ \bibinfo
  {author} {\bibfnamefont {Pingkun}\ \bibnamefont {Yan}},\ }\bibfield  {title}
  {\enquote {\bibinfo {title} {Adaptive shape prior constrained level sets for
  bladder {MR} image segmentation},}\ }\href@noop {} {\bibfield  {journal}
  {\bibinfo  {journal} {IEEE journal of biomedical and health informatics}\
  }\textbf {\bibinfo {volume} {18}},\ \bibinfo {pages} {1707--1716} (\bibinfo
  {year} {2014}{\natexlab{a}})}\BibitemShut {NoStop}%
\bibitem [{\citenamefont {Xu}\ \emph {et~al.}(2017{\natexlab{a}})\citenamefont
  {Xu}, \citenamefont {Zhang}, \citenamefont {Tian}, \citenamefont {Zhang},
  \citenamefont {Liu}, \citenamefont {Cui}, \citenamefont {Meng}, \citenamefont
  {Wu}, \citenamefont {Liu}, \citenamefont {Yang},\ and\ \citenamefont
  {Lu}}]{Xu2017}%
  \BibitemOpen
  \bibfield  {author} {\bibinfo {author} {\bibfnamefont {Xiaopan}\ \bibnamefont
  {Xu}}, \bibinfo {author} {\bibfnamefont {Xi}~\bibnamefont {Zhang}}, \bibinfo
  {author} {\bibfnamefont {Qiang}\ \bibnamefont {Tian}}, \bibinfo {author}
  {\bibfnamefont {Guopeng}\ \bibnamefont {Zhang}}, \bibinfo {author}
  {\bibfnamefont {Yang}\ \bibnamefont {Liu}}, \bibinfo {author} {\bibfnamefont
  {Guangbin}\ \bibnamefont {Cui}}, \bibinfo {author} {\bibfnamefont {Jiang}\
  \bibnamefont {Meng}}, \bibinfo {author} {\bibfnamefont {Yuxia}\ \bibnamefont
  {Wu}}, \bibinfo {author} {\bibfnamefont {Tianshuai}\ \bibnamefont {Liu}},
  \bibinfo {author} {\bibfnamefont {Zengyue}\ \bibnamefont {Yang}}, \ and\
  \bibinfo {author} {\bibfnamefont {Hongbing}\ \bibnamefont {Lu}},\ }\bibfield
  {title} {\enquote {\bibinfo {title} {{Three-dimensional texture features from
  intensity and high-order derivative maps for the discrimination between
  bladder tumors and wall tissues via MRI}},}\ }\href {\doibase
  10.1007/s11548-017-1522-8} {\bibfield  {journal} {\bibinfo  {journal}
  {International Journal of Computer Assisted Radiology and Surgery}\ }
  (\bibinfo {year} {2017}{\natexlab{a}}),\
  10.1007/s11548-017-1522-8}\BibitemShut {NoStop}%
\bibitem [{\citenamefont {Xu}\ \emph {et~al.}(2017{\natexlab{b}})\citenamefont
  {Xu}, \citenamefont {Liu}, \citenamefont {Zhang}, \citenamefont {Tian},
  \citenamefont {Wu}, \citenamefont {Zhang}, \citenamefont {Meng},
  \citenamefont {Yang},\ and\ \citenamefont {Lu}}]{Xu2017Preoperative}%
  \BibitemOpen
  \bibfield  {author} {\bibinfo {author} {\bibfnamefont {X.}~\bibnamefont
  {Xu}}, \bibinfo {author} {\bibfnamefont {Y.}~\bibnamefont {Liu}}, \bibinfo
  {author} {\bibfnamefont {X.}~\bibnamefont {Zhang}}, \bibinfo {author}
  {\bibfnamefont {Q.}~\bibnamefont {Tian}}, \bibinfo {author} {\bibfnamefont
  {Y.}~\bibnamefont {Wu}}, \bibinfo {author} {\bibfnamefont {G.}~\bibnamefont
  {Zhang}}, \bibinfo {author} {\bibfnamefont {J.}~\bibnamefont {Meng}},
  \bibinfo {author} {\bibfnamefont {Z.}~\bibnamefont {Yang}}, \ and\ \bibinfo
  {author} {\bibfnamefont {H.}~\bibnamefont {Lu}},\ }\bibfield  {title}
  {\enquote {\bibinfo {title} {Preoperative prediction of muscular invasiveness
  of bladder cancer with radiomic features on conventional {MRI} and its
  high-order derivative maps.}}\ }\href@noop {} {\bibfield  {journal} {\bibinfo
   {journal} {Abdominal Radiology}\ }\textbf {\bibinfo {volume} {42}},\
  \bibinfo {pages} {1--10} (\bibinfo {year} {2017}{\natexlab{b}})}\BibitemShut
  {NoStop}%
\bibitem [{\citenamefont {Zhang}\ \emph {et~al.}(2017)\citenamefont {Zhang},
  \citenamefont {Xu}, \citenamefont {Tian}, \citenamefont {Li}, \citenamefont
  {Wu}, \citenamefont {Yang}, \citenamefont {Liang}, \citenamefont {Liu},
  \citenamefont {Cui},\ and\ \citenamefont {Lu}}]{Zhang2017Radiomics}%
  \BibitemOpen
  \bibfield  {author} {\bibinfo {author} {\bibfnamefont {X.}~\bibnamefont
  {Zhang}}, \bibinfo {author} {\bibfnamefont {X.}~\bibnamefont {Xu}}, \bibinfo
  {author} {\bibfnamefont {Q.}~\bibnamefont {Tian}}, \bibinfo {author}
  {\bibfnamefont {B.}~\bibnamefont {Li}}, \bibinfo {author} {\bibfnamefont
  {Y.}~\bibnamefont {Wu}}, \bibinfo {author} {\bibfnamefont {Z.}~\bibnamefont
  {Yang}}, \bibinfo {author} {\bibfnamefont {Z.}~\bibnamefont {Liang}},
  \bibinfo {author} {\bibfnamefont {Y.}~\bibnamefont {Liu}}, \bibinfo {author}
  {\bibfnamefont {G.}~\bibnamefont {Cui}}, \ and\ \bibinfo {author}
  {\bibfnamefont {H.}~\bibnamefont {Lu}},\ }\bibfield  {title} {\enquote
  {\bibinfo {title} {Radiomics assessment of bladder cancer grade using texture
  features from diffusion-weighted imaging.}}\ }\href@noop {} {\bibfield
  {journal} {\bibinfo  {journal} {Journal of Magnetic Resonance Imaging Jmri}\
  }\textbf {\bibinfo {volume} {46}} (\bibinfo {year} {2017})}\BibitemShut
  {NoStop}%
\bibitem [{\citenamefont {Xiao}\ \emph {et~al.}(2016)\citenamefont {Xiao},
  \citenamefont {Zhang}, \citenamefont {Liu}, \citenamefont {Yang},
  \citenamefont {Zhang}, \citenamefont {Li}, \citenamefont {Jiao},\ and\
  \citenamefont {Lu}}]{xiao20163d}%
  \BibitemOpen
  \bibfield  {author} {\bibinfo {author} {\bibfnamefont {Dan}\ \bibnamefont
  {Xiao}}, \bibinfo {author} {\bibfnamefont {Guopeng}\ \bibnamefont {Zhang}},
  \bibinfo {author} {\bibfnamefont {Yang}\ \bibnamefont {Liu}}, \bibinfo
  {author} {\bibfnamefont {Zengyu}\ \bibnamefont {Yang}}, \bibinfo {author}
  {\bibfnamefont {Xi}~\bibnamefont {Zhang}}, \bibinfo {author} {\bibfnamefont
  {Lihong}\ \bibnamefont {Li}}, \bibinfo {author} {\bibfnamefont {Chun}\
  \bibnamefont {Jiao}}, \ and\ \bibinfo {author} {\bibfnamefont {Hongbing}\
  \bibnamefont {Lu}},\ }\bibfield  {title} {\enquote {\bibinfo {title} {3{D}
  detection and extraction of bladder tumors via {MR} virtual cystoscopy},}\
  }\href@noop {} {\bibfield  {journal} {\bibinfo  {journal} {International
  journal of computer assisted radiology and surgery}\ }\textbf {\bibinfo
  {volume} {11}},\ \bibinfo {pages} {89--97} (\bibinfo {year}
  {2016})}\BibitemShut {NoStop}%
\bibitem [{\citenamefont {Duan}\ \emph {et~al.}(2012)\citenamefont {Duan},
  \citenamefont {Yuan}, \citenamefont {Liu}, \citenamefont {Xiao},
  \citenamefont {Lv},\ and\ \citenamefont {Liang}}]{duan2012adaptive}%
  \BibitemOpen
  \bibfield  {author} {\bibinfo {author} {\bibfnamefont {Chaijie}\ \bibnamefont
  {Duan}}, \bibinfo {author} {\bibfnamefont {Kehong}\ \bibnamefont {Yuan}},
  \bibinfo {author} {\bibfnamefont {Fanghua}\ \bibnamefont {Liu}}, \bibinfo
  {author} {\bibfnamefont {Ping}\ \bibnamefont {Xiao}}, \bibinfo {author}
  {\bibfnamefont {Guoqing}\ \bibnamefont {Lv}}, \ and\ \bibinfo {author}
  {\bibfnamefont {Zhengrong}\ \bibnamefont {Liang}},\ }\bibfield  {title}
  {\enquote {\bibinfo {title} {An adaptive window-setting scheme for
  segmentation of bladder tumor surface via {MR} cystography},}\ }\href@noop {}
  {\bibfield  {journal} {\bibinfo  {journal} {IEEE Transactions on Information
  Technology in Biomedicine}\ }\textbf {\bibinfo {volume} {16}},\ \bibinfo
  {pages} {720--729} (\bibinfo {year} {2012})}\BibitemShut {NoStop}%
\bibitem [{\citenamefont {Lambin}\ \emph {et~al.}(2017)\citenamefont {Lambin},
  \citenamefont {Leijenaar}, \citenamefont {Deist}, \citenamefont {Peerlings},
  \citenamefont {Jong}, \citenamefont {Timmeren}, \citenamefont {Sanduleanu},
  \citenamefont {Larue}, \citenamefont {Even},\ and\ \citenamefont
  {Jochems}}]{Lambin2017Radiomics}%
  \BibitemOpen
  \bibfield  {author} {\bibinfo {author} {\bibfnamefont {Philippe}\
  \bibnamefont {Lambin}}, \bibinfo {author} {\bibfnamefont {Ralph T.~H.}\
  \bibnamefont {Leijenaar}}, \bibinfo {author} {\bibfnamefont {Timo~M.}\
  \bibnamefont {Deist}}, \bibinfo {author} {\bibfnamefont {Jurgen}\
  \bibnamefont {Peerlings}}, \bibinfo {author} {\bibfnamefont {Evelyn E.
  C.~De}\ \bibnamefont {Jong}}, \bibinfo {author} {\bibfnamefont {Janita~Van}\
  \bibnamefont {Timmeren}}, \bibinfo {author} {\bibfnamefont {Sebastian}\
  \bibnamefont {Sanduleanu}}, \bibinfo {author} {\bibfnamefont {Ruben T.
  H.~M.}\ \bibnamefont {Larue}}, \bibinfo {author} {\bibfnamefont {Aniek
  J.~G.}\ \bibnamefont {Even}}, \ and\ \bibinfo {author} {\bibfnamefont
  {Arthur}\ \bibnamefont {Jochems}},\ }\bibfield  {title} {\enquote {\bibinfo
  {title} {Radiomics: the bridge between medical imaging and personalized
  medicine},}\ }\href@noop {} {\bibfield  {journal} {\bibinfo  {journal}
  {Nature Reviews Clinical Oncology}\ }\textbf {\bibinfo {volume} {14}},\
  \bibinfo {pages} {749} (\bibinfo {year} {2017})}\BibitemShut {NoStop}%
\bibitem [{\citenamefont {Xu}\ \emph {et~al.}(2017{\natexlab{c}})\citenamefont
  {Xu}, \citenamefont {Zhang}, \citenamefont {Liu}, \citenamefont {Tian},
  \citenamefont {Zhang}, \citenamefont {Yang}, \citenamefont {Lu},\ and\
  \citenamefont {Yuan}}]{xu2017simultaneous}%
  \BibitemOpen
  \bibfield  {author} {\bibinfo {author} {\bibfnamefont {Xiaopan}\ \bibnamefont
  {Xu}}, \bibinfo {author} {\bibfnamefont {Xi}~\bibnamefont {Zhang}}, \bibinfo
  {author} {\bibfnamefont {Yang}\ \bibnamefont {Liu}}, \bibinfo {author}
  {\bibfnamefont {Qiang}\ \bibnamefont {Tian}}, \bibinfo {author}
  {\bibfnamefont {Guopeng}\ \bibnamefont {Zhang}}, \bibinfo {author}
  {\bibfnamefont {Zengyue}\ \bibnamefont {Yang}}, \bibinfo {author}
  {\bibfnamefont {Hongbing}\ \bibnamefont {Lu}}, \ and\ \bibinfo {author}
  {\bibfnamefont {Jing}\ \bibnamefont {Yuan}},\ }\bibfield  {title} {\enquote
  {\bibinfo {title} {Simultaneous segmentation of multiple regions in {3D}
  bladder {MRI} by efficient convex optimization of coupled surfaces},}\ }in\
  \href@noop {} {\emph {\bibinfo {booktitle} {International Conference on Image
  and Graphics}}}\ (\bibinfo {organization} {Springer},\ \bibinfo {year}
  {2017})\ pp.\ \bibinfo {pages} {528--542}\BibitemShut {NoStop}%
\bibitem [{\citenamefont {Li}\ \emph {et~al.}(2004)\citenamefont {Li},
  \citenamefont {Wang}, \citenamefont {Li}, \citenamefont {Wei}, \citenamefont
  {Adler}, \citenamefont {Huang}, \citenamefont {Rizvi}, \citenamefont {Meng},
  \citenamefont {Harrington},\ and\ \citenamefont {Liang}}]{li2004new}%
  \BibitemOpen
  \bibfield  {author} {\bibinfo {author} {\bibfnamefont {Lihong}\ \bibnamefont
  {Li}}, \bibinfo {author} {\bibfnamefont {Zigang}\ \bibnamefont {Wang}},
  \bibinfo {author} {\bibfnamefont {Xiang}\ \bibnamefont {Li}}, \bibinfo
  {author} {\bibfnamefont {Xinzhou}\ \bibnamefont {Wei}}, \bibinfo {author}
  {\bibfnamefont {Howard~L}\ \bibnamefont {Adler}}, \bibinfo {author}
  {\bibfnamefont {Wei}\ \bibnamefont {Huang}}, \bibinfo {author} {\bibfnamefont
  {Syed~A}\ \bibnamefont {Rizvi}}, \bibinfo {author} {\bibfnamefont {Hong}\
  \bibnamefont {Meng}}, \bibinfo {author} {\bibfnamefont {Donald~P}\
  \bibnamefont {Harrington}}, \ and\ \bibinfo {author} {\bibfnamefont
  {Zhengrong}\ \bibnamefont {Liang}},\ }\bibfield  {title} {\enquote {\bibinfo
  {title} {A new partial volume segmentation approach to extract bladder wall
  for computer-aided detection in virtual cystoscopy},}\ }in\ \href@noop {}
  {\emph {\bibinfo {booktitle} {Medical Imaging 2004: Physiology, Function, and
  Structure from Medical Images}}},\ Vol.\ \bibinfo {volume} {5369}\ (\bibinfo
  {organization} {International Society for Optics and Photonics},\ \bibinfo
  {year} {2004})\ pp.\ \bibinfo {pages} {199--207}\BibitemShut {NoStop}%
\bibitem [{\citenamefont {Li}\ \emph {et~al.}(2008)\citenamefont {Li},
  \citenamefont {Liang}, \citenamefont {Wang}, \citenamefont {Lu},
  \citenamefont {Wei}, \citenamefont {Wagshul}, \citenamefont {Zawin},
  \citenamefont {Posniak},\ and\ \citenamefont {Lee}}]{li2008segmentation}%
  \BibitemOpen
  \bibfield  {author} {\bibinfo {author} {\bibfnamefont {Lihong}\ \bibnamefont
  {Li}}, \bibinfo {author} {\bibfnamefont {Zhengrong}\ \bibnamefont {Liang}},
  \bibinfo {author} {\bibfnamefont {Su}~\bibnamefont {Wang}}, \bibinfo {author}
  {\bibfnamefont {Hongyu}\ \bibnamefont {Lu}}, \bibinfo {author} {\bibfnamefont
  {Xinzhou}\ \bibnamefont {Wei}}, \bibinfo {author} {\bibfnamefont {Mark}\
  \bibnamefont {Wagshul}}, \bibinfo {author} {\bibfnamefont {Marlene}\
  \bibnamefont {Zawin}}, \bibinfo {author} {\bibfnamefont {Erica~J}\
  \bibnamefont {Posniak}}, \ and\ \bibinfo {author} {\bibfnamefont
  {Christopher~S}\ \bibnamefont {Lee}},\ }\bibfield  {title} {\enquote
  {\bibinfo {title} {Segmentation of multispectral bladder {MR} images with
  inhomogeneity correction for virtual cystoscopy},}\ }in\ \href@noop {} {\emph
  {\bibinfo {booktitle} {Medical Imaging 2008: Physiology, Function, and
  Structure from Medical Images}}},\ Vol.\ \bibinfo {volume} {6916}\ (\bibinfo
  {organization} {International Society for Optics and Photonics},\ \bibinfo
  {year} {2008})\ p.\ \bibinfo {pages} {69160U}\BibitemShut {NoStop}%
\bibitem [{\citenamefont {Chi}\ \emph {et~al.}(2011)\citenamefont {Chi},
  \citenamefont {Brady}, \citenamefont {Moore},\ and\ \citenamefont
  {Schnabel}}]{chi2011segmentation}%
  \BibitemOpen
  \bibfield  {author} {\bibinfo {author} {\bibfnamefont {J~Wenjun}\
  \bibnamefont {Chi}}, \bibinfo {author} {\bibfnamefont {Michael}\ \bibnamefont
  {Brady}}, \bibinfo {author} {\bibfnamefont {Niall~R}\ \bibnamefont {Moore}},
  \ and\ \bibinfo {author} {\bibfnamefont {Julia~A}\ \bibnamefont {Schnabel}},\
  }\bibfield  {title} {\enquote {\bibinfo {title} {Segmentation of the bladder
  wall using coupled level set methods},}\ }in\ \href@noop {} {\emph {\bibinfo
  {booktitle} {Biomedical Imaging: From Nano to Macro, 2011 IEEE International
  Symposium on}}}\ (\bibinfo {organization} {IEEE},\ \bibinfo {year} {2011})\
  pp.\ \bibinfo {pages} {1653--1656}\BibitemShut {NoStop}%
\bibitem [{\citenamefont {Garnier}\ \emph {et~al.}(2011)\citenamefont
  {Garnier}, \citenamefont {Ke},\ and\ \citenamefont
  {Dillenseger}}]{garnier2011bladder}%
  \BibitemOpen
  \bibfield  {author} {\bibinfo {author} {\bibfnamefont {Carole}\ \bibnamefont
  {Garnier}}, \bibinfo {author} {\bibfnamefont {Wu}~\bibnamefont {Ke}}, \ and\
  \bibinfo {author} {\bibfnamefont {Jean-Louis}\ \bibnamefont {Dillenseger}},\
  }\bibfield  {title} {\enquote {\bibinfo {title} {Bladder segmentation in
  {MRI} images using active region growing model},}\ }in\ \href@noop {} {\emph
  {\bibinfo {booktitle} {Engineering in Medicine and Biology Society, EMBC,
  2011 Annual International Conference of the IEEE}}}\ (\bibinfo {organization}
  {IEEE},\ \bibinfo {year} {2011})\ pp.\ \bibinfo {pages}
  {5702--5705}\BibitemShut {NoStop}%
\bibitem [{\citenamefont {Ma}\ \emph {et~al.}(2011)\citenamefont {Ma},
  \citenamefont {Jorge}, \citenamefont {Mascarenhas},\ and\ \citenamefont
  {Tavares}}]{ma2011novel}%
  \BibitemOpen
  \bibfield  {author} {\bibinfo {author} {\bibfnamefont {Zhen}\ \bibnamefont
  {Ma}}, \bibinfo {author} {\bibfnamefont {Renato~Natal}\ \bibnamefont
  {Jorge}}, \bibinfo {author} {\bibfnamefont {T}~\bibnamefont {Mascarenhas}}, \
  and\ \bibinfo {author} {\bibfnamefont {Jo{\~a}o Manuel~RS}\ \bibnamefont
  {Tavares}},\ }\bibfield  {title} {\enquote {\bibinfo {title} {Novel approach
  to segment the inner and outer boundaries of the bladder wall in
  {T2}-weighted magnetic resonance images},}\ }\href@noop {} {\bibfield
  {journal} {\bibinfo  {journal} {Annals of biomedical engineering}\ }\textbf
  {\bibinfo {volume} {39}},\ \bibinfo {pages} {2287--2297} (\bibinfo {year}
  {2011})}\BibitemShut {NoStop}%
\bibitem [{\citenamefont {Han}\ \emph {et~al.}(2013)\citenamefont {Han},
  \citenamefont {Li}, \citenamefont {Duan}, \citenamefont {Zhang},
  \citenamefont {Zhao},\ and\ \citenamefont {Liang}}]{han2013unified}%
  \BibitemOpen
  \bibfield  {author} {\bibinfo {author} {\bibfnamefont {Hao}\ \bibnamefont
  {Han}}, \bibinfo {author} {\bibfnamefont {Lihong}\ \bibnamefont {Li}},
  \bibinfo {author} {\bibfnamefont {Chaijie}\ \bibnamefont {Duan}}, \bibinfo
  {author} {\bibfnamefont {Hao}\ \bibnamefont {Zhang}}, \bibinfo {author}
  {\bibfnamefont {Yang}\ \bibnamefont {Zhao}}, \ and\ \bibinfo {author}
  {\bibfnamefont {Zhengrong}\ \bibnamefont {Liang}},\ }\bibfield  {title}
  {\enquote {\bibinfo {title} {A unified {EM} approach to bladder wall
  segmentation with coupled level-set constraints},}\ }\href@noop {} {\bibfield
   {journal} {\bibinfo  {journal} {Medical image analysis}\ }\textbf {\bibinfo
  {volume} {17}},\ \bibinfo {pages} {1192--1205} (\bibinfo {year}
  {2013})}\BibitemShut {NoStop}%
\bibitem [{\citenamefont {Qin}\ \emph {et~al.}(2014{\natexlab{b}})\citenamefont
  {Qin}, \citenamefont {Li}, \citenamefont {Liu}, \citenamefont {Lu},\ and\
  \citenamefont {Yan}}]{Qin2014}%
  \BibitemOpen
  \bibfield  {author} {\bibinfo {author} {\bibfnamefont {Xianjing}\
  \bibnamefont {Qin}}, \bibinfo {author} {\bibfnamefont {Xuelong}\ \bibnamefont
  {Li}}, \bibinfo {author} {\bibfnamefont {Yang}\ \bibnamefont {Liu}}, \bibinfo
  {author} {\bibfnamefont {Hongbing}\ \bibnamefont {Lu}}, \ and\ \bibinfo
  {author} {\bibfnamefont {Pingkun}\ \bibnamefont {Yan}},\ }\bibfield  {title}
  {\enquote {\bibinfo {title} {{Adaptive Shape Prior Constrained Level Sets for
  Bladder MR Image Segmentation}},}\ }\href {\doibase
  10.1109/JBHI.2013.2288935} {\bibfield  {journal} {\bibinfo  {journal} {IEEE
  Journal of Biomedical and Health Informatics}\ }\textbf {\bibinfo {volume}
  {18}},\ \bibinfo {pages} {1707--1716} (\bibinfo {year}
  {2014}{\natexlab{b}})}\BibitemShut {NoStop}%
\bibitem [{\citenamefont {Bezdek}\ \emph {et~al.}(1984)\citenamefont {Bezdek},
  \citenamefont {Ehrlich},\ and\ \citenamefont {Full}}]{bezdek1984fcm}%
  \BibitemOpen
  \bibfield  {author} {\bibinfo {author} {\bibfnamefont {James~C}\ \bibnamefont
  {Bezdek}}, \bibinfo {author} {\bibfnamefont {Robert}\ \bibnamefont
  {Ehrlich}}, \ and\ \bibinfo {author} {\bibfnamefont {William}\ \bibnamefont
  {Full}},\ }\bibfield  {title} {\enquote {\bibinfo {title} {{FCM}: The fuzzy
  c-means clustering algorithm},}\ }\href@noop {} {\bibfield  {journal}
  {\bibinfo  {journal} {Computers \& Geosciences}\ }\textbf {\bibinfo {volume}
  {10}},\ \bibinfo {pages} {191--203} (\bibinfo {year} {1984})}\BibitemShut
  {NoStop}%
\bibitem [{\citenamefont {Huang}\ \emph {et~al.}(2017)\citenamefont {Huang},
  \citenamefont {Liu}, \citenamefont {Weinberger},\ and\ \citenamefont {van~der
  Maaten}}]{huang2017densely}%
  \BibitemOpen
  \bibfield  {author} {\bibinfo {author} {\bibfnamefont {Gao}\ \bibnamefont
  {Huang}}, \bibinfo {author} {\bibfnamefont {Zhuang}\ \bibnamefont {Liu}},
  \bibinfo {author} {\bibfnamefont {Kilian~Q}\ \bibnamefont {Weinberger}}, \
  and\ \bibinfo {author} {\bibfnamefont {Laurens}\ \bibnamefont {van~der
  Maaten}},\ }\bibfield  {title} {\enquote {\bibinfo {title} {Densely connected
  convolutional networks},}\ }in\ \href@noop {} {\emph {\bibinfo {booktitle}
  {Proceedings of the IEEE conference on computer vision and pattern
  recognition}}},\ Vol.~\bibinfo {volume} {1}\ (\bibinfo {year} {2017})\
  p.~\bibinfo {pages} {3}\BibitemShut {NoStop}%
\bibitem [{\citenamefont {Redmon}\ and\ \citenamefont
  {Farhadi}(2017)}]{redmon2016yolo9000}%
  \BibitemOpen
  \bibfield  {author} {\bibinfo {author} {\bibfnamefont {Joseph}\ \bibnamefont
  {Redmon}}\ and\ \bibinfo {author} {\bibfnamefont {Ali}\ \bibnamefont
  {Farhadi}},\ }\bibfield  {title} {\enquote {\bibinfo {title} {{YOLO9000}:
  better, faster, stronger},}\ }\bibfield  {booktitle} {\emph {\bibinfo
  {booktitle} {Proceedings of the IEEE Conference on Computer Vision and
  Pattern Recognition}},\ }\href@noop {} {\ ,\ \bibinfo {pages} {7263--7271}
  (\bibinfo {year} {2017})}\BibitemShut {NoStop}%
\bibitem [{\citenamefont {Yu}\ and\ \citenamefont
  {Koltun}(2015)}]{yu2015multi}%
  \BibitemOpen
  \bibfield  {author} {\bibinfo {author} {\bibfnamefont {Fisher}\ \bibnamefont
  {Yu}}\ and\ \bibinfo {author} {\bibfnamefont {Vladlen}\ \bibnamefont
  {Koltun}},\ }\bibfield  {title} {\enquote {\bibinfo {title} {Multi-scale
  context aggregation by dilated convolutions},}\ }\href@noop {} {\bibfield
  {journal} {\bibinfo  {journal} {arXiv preprint arXiv:1511.07122}\ } (\bibinfo
  {year} {2015})}\BibitemShut {NoStop}%
\bibitem [{\citenamefont {Dolz}\ \emph
  {et~al.}(2018{\natexlab{a}})\citenamefont {Dolz}, \citenamefont
  {Desrosiers},\ and\ \citenamefont {Ayed}}]{DolzNeuro2017}%
  \BibitemOpen
  \bibfield  {author} {\bibinfo {author} {\bibfnamefont {J.}~\bibnamefont
  {Dolz}}, \bibinfo {author} {\bibfnamefont {C.}~\bibnamefont {Desrosiers}}, \
  and\ \bibinfo {author} {\bibfnamefont {I.~Ben}\ \bibnamefont {Ayed}},\
  }\bibfield  {title} {\enquote {\bibinfo {title} {{3D} fully convolutional
  networks for subcortical segmentation in {MRI}: A large-scale study},}\
  }\href@noop {} {\bibfield  {journal} {\bibinfo  {journal} {NeuroImage}\
  }\textbf {\bibinfo {volume} {170}},\ \bibinfo {pages} {456--470} (\bibinfo
  {year} {2018}{\natexlab{a}})}\BibitemShut {NoStop}%
\bibitem [{\citenamefont {Fechter}\ \emph {et~al.}(2017)\citenamefont
  {Fechter}, \citenamefont {Adebahr}, \citenamefont {Baltas}, \citenamefont
  {Ben~Ayed}, \citenamefont {Desrosiers},\ and\ \citenamefont
  {Dolz}}]{Fechter_Esophagus}%
  \BibitemOpen
  \bibfield  {author} {\bibinfo {author} {\bibfnamefont {Tobias}\ \bibnamefont
  {Fechter}}, \bibinfo {author} {\bibfnamefont {Sonja}\ \bibnamefont
  {Adebahr}}, \bibinfo {author} {\bibfnamefont {Dimos}\ \bibnamefont {Baltas}},
  \bibinfo {author} {\bibfnamefont {Ismail}\ \bibnamefont {Ben~Ayed}}, \bibinfo
  {author} {\bibfnamefont {Christian}\ \bibnamefont {Desrosiers}}, \ and\
  \bibinfo {author} {\bibfnamefont {Jose}\ \bibnamefont {Dolz}},\ }\bibfield
  {title} {\enquote {\bibinfo {title} {Esophagus segmentation in {CT} via {3D}
  fully convolutional neural network and random walk},}\ }\href@noop {}
  {\bibfield  {journal} {\bibinfo  {journal} {Medical Physics}\ }\textbf
  {\bibinfo {volume} {44}},\ \bibinfo {pages} {6341--6352} (\bibinfo {year}
  {2017})}\BibitemShut {NoStop}%
\bibitem [{\citenamefont {Litjens}\ \emph {et~al.}(2017)\citenamefont
  {Litjens}, \citenamefont {Kooi}, \citenamefont {Bejnordi}, \citenamefont
  {Setio}, \citenamefont {Ciompi}, \citenamefont {Ghafoorian}, \citenamefont
  {van~der Laak}, \citenamefont {Van~Ginneken},\ and\ \citenamefont
  {S{\'a}nchez}}]{litjens2017survey}%
  \BibitemOpen
  \bibfield  {author} {\bibinfo {author} {\bibfnamefont {Geert}\ \bibnamefont
  {Litjens}}, \bibinfo {author} {\bibfnamefont {Thijs}\ \bibnamefont {Kooi}},
  \bibinfo {author} {\bibfnamefont {Babak~Ehteshami}\ \bibnamefont {Bejnordi}},
  \bibinfo {author} {\bibfnamefont {Arnaud Arindra~Adiyoso}\ \bibnamefont
  {Setio}}, \bibinfo {author} {\bibfnamefont {Francesco}\ \bibnamefont
  {Ciompi}}, \bibinfo {author} {\bibfnamefont {Mohsen}\ \bibnamefont
  {Ghafoorian}}, \bibinfo {author} {\bibfnamefont {Jeroen~AWM}\ \bibnamefont
  {van~der Laak}}, \bibinfo {author} {\bibfnamefont {Bram}\ \bibnamefont
  {Van~Ginneken}}, \ and\ \bibinfo {author} {\bibfnamefont {Clara~I}\
  \bibnamefont {S{\'a}nchez}},\ }\bibfield  {title} {\enquote {\bibinfo {title}
  {A survey on deep learning in medical image analysis},}\ }\href@noop {}
  {\bibfield  {journal} {\bibinfo  {journal} {Medical image analysis}\ }\textbf
  {\bibinfo {volume} {42}},\ \bibinfo {pages} {60--88} (\bibinfo {year}
  {2017})}\BibitemShut {NoStop}%
\bibitem [{\citenamefont {Dolz}\ \emph
  {et~al.}(2018{\natexlab{b}})\citenamefont {Dolz}, \citenamefont {Gopinath},
  \citenamefont {Yuan}, \citenamefont {Lombaert}, \citenamefont {Desrosiers},\
  and\ \citenamefont {Ben~Ayed}}]{dolz2018hyperdense}%
  \BibitemOpen
  \bibfield  {author} {\bibinfo {author} {\bibfnamefont {Jose}\ \bibnamefont
  {Dolz}}, \bibinfo {author} {\bibfnamefont {Karthik}\ \bibnamefont
  {Gopinath}}, \bibinfo {author} {\bibfnamefont {Jing}\ \bibnamefont {Yuan}},
  \bibinfo {author} {\bibfnamefont {Herve}\ \bibnamefont {Lombaert}}, \bibinfo
  {author} {\bibfnamefont {Christian}\ \bibnamefont {Desrosiers}}, \ and\
  \bibinfo {author} {\bibfnamefont {Ismail}\ \bibnamefont {Ben~Ayed}},\
  }\bibfield  {title} {\enquote {\bibinfo {title} {Hyperdense-{N}et: {A}
  hyper-densely connected {CNN} for multi-modal image segmentation},}\
  }\href@noop {} {\bibfield  {journal} {\bibinfo  {journal} {arXiv preprint
  arXiv:1804.02967}\ } (\bibinfo {year} {2018}{\natexlab{b}})}\BibitemShut
  {NoStop}%
\bibitem [{\citenamefont {Carass}\ \emph {et~al.}(2018)\citenamefont {Carass},
  \citenamefont {Cuzzocreo}, \citenamefont {Han}, \citenamefont
  {Hernandez-Castillo}, \citenamefont {Rasser}, \citenamefont {Ganz},
  \citenamefont {Beliveau}, \citenamefont {Dolz}, \citenamefont {Ayed},
  \citenamefont {Desrosiers} \emph {et~al.}}]{carass2018comparing}%
  \BibitemOpen
  \bibfield  {author} {\bibinfo {author} {\bibfnamefont {Aaron}\ \bibnamefont
  {Carass}}, \bibinfo {author} {\bibfnamefont {Jennifer~L}\ \bibnamefont
  {Cuzzocreo}}, \bibinfo {author} {\bibfnamefont {Shuo}\ \bibnamefont {Han}},
  \bibinfo {author} {\bibfnamefont {Carlos~R}\ \bibnamefont
  {Hernandez-Castillo}}, \bibinfo {author} {\bibfnamefont {Paul~E}\
  \bibnamefont {Rasser}}, \bibinfo {author} {\bibfnamefont {Melanie}\
  \bibnamefont {Ganz}}, \bibinfo {author} {\bibfnamefont {Vincent}\
  \bibnamefont {Beliveau}}, \bibinfo {author} {\bibfnamefont {Jose}\
  \bibnamefont {Dolz}}, \bibinfo {author} {\bibfnamefont {Ismail~Ben}\
  \bibnamefont {Ayed}}, \bibinfo {author} {\bibfnamefont {Christian}\
  \bibnamefont {Desrosiers}},  \emph {et~al.},\ }\bibfield  {title} {\enquote
  {\bibinfo {title} {Comparing fully automated state-of-the-art cerebellum
  parcellation from magnetic resonance images},}\ }\href@noop {} {\bibfield
  {journal} {\bibinfo  {journal} {NeuroImage}\ }\textbf {\bibinfo {volume}
  {183}},\ \bibinfo {pages} {150--172} (\bibinfo {year} {2018})}\BibitemShut
  {NoStop}%
\bibitem [{\citenamefont {Cha}\ \emph {et~al.}(2016{\natexlab{a}})\citenamefont
  {Cha}, \citenamefont {Hadjiiski}, \citenamefont {Samala}, \citenamefont
  {Chan}, \citenamefont {Caoili},\ and\ \citenamefont
  {Cohan}}]{cha2016urinary}%
  \BibitemOpen
  \bibfield  {author} {\bibinfo {author} {\bibfnamefont {Kenny~H}\ \bibnamefont
  {Cha}}, \bibinfo {author} {\bibfnamefont {Lubomir}\ \bibnamefont
  {Hadjiiski}}, \bibinfo {author} {\bibfnamefont {Ravi~K}\ \bibnamefont
  {Samala}}, \bibinfo {author} {\bibfnamefont {Heang-Ping}\ \bibnamefont
  {Chan}}, \bibinfo {author} {\bibfnamefont {Elaine~M}\ \bibnamefont {Caoili}},
  \ and\ \bibinfo {author} {\bibfnamefont {Richard~H}\ \bibnamefont {Cohan}},\
  }\bibfield  {title} {\enquote {\bibinfo {title} {Urinary bladder segmentation
  in {CT} urography using deep-learning convolutional neural network and level
  sets},}\ }\href@noop {} {\bibfield  {journal} {\bibinfo  {journal} {Medical
  physics}\ }\textbf {\bibinfo {volume} {43}},\ \bibinfo {pages} {1882--1896}
  (\bibinfo {year} {2016}{\natexlab{a}})}\BibitemShut {NoStop}%
\bibitem [{\citenamefont {Cha}\ \emph {et~al.}(2016{\natexlab{b}})\citenamefont
  {Cha}, \citenamefont {Hadjiiski}, \citenamefont {Samala}, \citenamefont
  {Chan}, \citenamefont {Cohan}, \citenamefont {Caoili}, \citenamefont
  {Paramagul}, \citenamefont {Alva},\ and\ \citenamefont
  {Weizer}}]{cha2016bladder}%
  \BibitemOpen
  \bibfield  {author} {\bibinfo {author} {\bibfnamefont {Kenny~H}\ \bibnamefont
  {Cha}}, \bibinfo {author} {\bibfnamefont {Lubomir~M}\ \bibnamefont
  {Hadjiiski}}, \bibinfo {author} {\bibfnamefont {Ravi~K}\ \bibnamefont
  {Samala}}, \bibinfo {author} {\bibfnamefont {Heang-Ping}\ \bibnamefont
  {Chan}}, \bibinfo {author} {\bibfnamefont {Richard~H}\ \bibnamefont {Cohan}},
  \bibinfo {author} {\bibfnamefont {Elaine~M}\ \bibnamefont {Caoili}}, \bibinfo
  {author} {\bibfnamefont {Chintana}\ \bibnamefont {Paramagul}}, \bibinfo
  {author} {\bibfnamefont {Ajjai}\ \bibnamefont {Alva}}, \ and\ \bibinfo
  {author} {\bibfnamefont {Alon~Z}\ \bibnamefont {Weizer}},\ }\bibfield
  {title} {\enquote {\bibinfo {title} {Bladder cancer segmentation in {CT} for
  treatment response assessment: application of deep-learning convolution
  neural network—a pilot study},}\ }\href@noop {} {\bibfield  {journal}
  {\bibinfo  {journal} {Tomography: a journal for imaging research}\ }\textbf
  {\bibinfo {volume} {2}},\ \bibinfo {pages} {421} (\bibinfo {year}
  {2016}{\natexlab{b}})}\BibitemShut {NoStop}%
\bibitem [{\citenamefont {Men}\ \emph {et~al.}(2017)\citenamefont {Men},
  \citenamefont {Dai},\ and\ \citenamefont {Li}}]{men2017automatic}%
  \BibitemOpen
  \bibfield  {author} {\bibinfo {author} {\bibfnamefont {Kuo}\ \bibnamefont
  {Men}}, \bibinfo {author} {\bibfnamefont {Jianrong}\ \bibnamefont {Dai}}, \
  and\ \bibinfo {author} {\bibfnamefont {Yexiong}\ \bibnamefont {Li}},\
  }\bibfield  {title} {\enquote {\bibinfo {title} {Automatic segmentation of
  the clinical target volume and organs at risk in the planning {CT} for rectal
  cancer using deep dilated convolutional neural networks},}\ }\href@noop {}
  {\bibfield  {journal} {\bibinfo  {journal} {Medical physics}\ }\textbf
  {\bibinfo {volume} {44}},\ \bibinfo {pages} {6377--6389} (\bibinfo {year}
  {2017})}\BibitemShut {NoStop}%
\bibitem [{\citenamefont {Ronneberger}\ \emph {et~al.}(2015)\citenamefont
  {Ronneberger}, \citenamefont {Fischer},\ and\ \citenamefont
  {Brox}}]{ronneberger2015u}%
  \BibitemOpen
  \bibfield  {author} {\bibinfo {author} {\bibfnamefont {Olaf}\ \bibnamefont
  {Ronneberger}}, \bibinfo {author} {\bibfnamefont {Philipp}\ \bibnamefont
  {Fischer}}, \ and\ \bibinfo {author} {\bibfnamefont {Thomas}\ \bibnamefont
  {Brox}},\ }\bibfield  {title} {\enquote {\bibinfo {title} {U-net:
  Convolutional networks for biomedical image segmentation},}\ }in\ \href@noop
  {} {\emph {\bibinfo {booktitle} {International Conference on Medical image
  computing and computer-assisted intervention}}}\ (\bibinfo {organization}
  {Springer},\ \bibinfo {year} {2015})\ pp.\ \bibinfo {pages}
  {234--241}\BibitemShut {NoStop}%
\bibitem [{\citenamefont {Hamaguchi}\ \emph {et~al.}(2018)\citenamefont
  {Hamaguchi}, \citenamefont {Fujita}, \citenamefont {Nemoto}, \citenamefont
  {Imaizumi},\ and\ \citenamefont {Hikosaka}}]{hamaguchi2018effective}%
  \BibitemOpen
  \bibfield  {author} {\bibinfo {author} {\bibfnamefont {Ryuhei}\ \bibnamefont
  {Hamaguchi}}, \bibinfo {author} {\bibfnamefont {Aito}\ \bibnamefont
  {Fujita}}, \bibinfo {author} {\bibfnamefont {Keisuke}\ \bibnamefont
  {Nemoto}}, \bibinfo {author} {\bibfnamefont {Tomoyuki}\ \bibnamefont
  {Imaizumi}}, \ and\ \bibinfo {author} {\bibfnamefont {Shuhei}\ \bibnamefont
  {Hikosaka}},\ }\bibfield  {title} {\enquote {\bibinfo {title} {Effective use
  of dilated convolutions for segmenting small object instances in remote
  sensing imagery},}\ }in\ \href@noop {} {\emph {\bibinfo {booktitle} {2018
  IEEE Winter Conference on Applications of Computer Vision (WACV)}}}\
  (\bibinfo {organization} {IEEE},\ \bibinfo {year} {2018})\ pp.\ \bibinfo
  {pages} {1442--1450}\BibitemShut {NoStop}%
\bibitem [{\citenamefont {Krizhevsky}\ \emph {et~al.}(2012)\citenamefont
  {Krizhevsky}, \citenamefont {Sutskever},\ and\ \citenamefont
  {Hinton}}]{krizhevsky2012imagenet}%
  \BibitemOpen
  \bibfield  {author} {\bibinfo {author} {\bibfnamefont {Alex}\ \bibnamefont
  {Krizhevsky}}, \bibinfo {author} {\bibfnamefont {Ilya}\ \bibnamefont
  {Sutskever}}, \ and\ \bibinfo {author} {\bibfnamefont {Geoffrey~E}\
  \bibnamefont {Hinton}},\ }\bibfield  {title} {\enquote {\bibinfo {title}
  {Imagenet classification with deep convolutional neural networks},}\ }in\
  \href@noop {} {\emph {\bibinfo {booktitle} {Advances in neural information
  processing systems}}}\ (\bibinfo {year} {2012})\ pp.\ \bibinfo {pages}
  {1097--1105}\BibitemShut {NoStop}%
\bibitem [{\citenamefont {LeCun}\ \emph {et~al.}(1998)\citenamefont {LeCun},
  \citenamefont {Bottou}, \citenamefont {Bengio},\ and\ \citenamefont
  {Haffner}}]{lecun1998gradient}%
  \BibitemOpen
  \bibfield  {author} {\bibinfo {author} {\bibfnamefont {Yann}\ \bibnamefont
  {LeCun}}, \bibinfo {author} {\bibfnamefont {L{\'e}on}\ \bibnamefont
  {Bottou}}, \bibinfo {author} {\bibfnamefont {Yoshua}\ \bibnamefont {Bengio}},
  \ and\ \bibinfo {author} {\bibfnamefont {Patrick}\ \bibnamefont {Haffner}},\
  }\bibfield  {title} {\enquote {\bibinfo {title} {Gradient-based learning
  applied to document recognition},}\ }\href@noop {} {\bibfield  {journal}
  {\bibinfo  {journal} {Proceedings of the IEEE}\ }\textbf {\bibinfo {volume}
  {86}},\ \bibinfo {pages} {2278--2324} (\bibinfo {year} {1998})}\BibitemShut
  {NoStop}%
\bibitem [{\citenamefont {Long}\ \emph {et~al.}(2015)\citenamefont {Long},
  \citenamefont {Shelhamer},\ and\ \citenamefont {Darrell}}]{FCN}%
  \BibitemOpen
  \bibfield  {author} {\bibinfo {author} {\bibfnamefont {Jonathan}\
  \bibnamefont {Long}}, \bibinfo {author} {\bibfnamefont {Evan}\ \bibnamefont
  {Shelhamer}}, \ and\ \bibinfo {author} {\bibfnamefont {Trevor}\ \bibnamefont
  {Darrell}},\ }\bibfield  {title} {\enquote {\bibinfo {title} {Fully
  convolutional networks for semantic segmentation},}\ }in\ \href@noop {}
  {\emph {\bibinfo {booktitle} {Proceedings of the IEEE Conference on Computer
  Vision and Pattern Recognition}}}\ (\bibinfo {year} {2015})\ pp.\ \bibinfo
  {pages} {3431--3440}\BibitemShut {NoStop}%
\bibitem [{\citenamefont {Christ}\ \emph {et~al.}(2016)\citenamefont {Christ},
  \citenamefont {Elshaer}, \citenamefont {Ettlinger}, \citenamefont
  {Tatavarty}, \citenamefont {Bickel}, \citenamefont {Bilic}, \citenamefont
  {Rempfler}, \citenamefont {Armbruster}, \citenamefont {Hofmann},
  \citenamefont {D’Anastasi} \emph {et~al.}}]{christ2016automatic}%
  \BibitemOpen
  \bibfield  {author} {\bibinfo {author} {\bibfnamefont {Patrick~Ferdinand}\
  \bibnamefont {Christ}}, \bibinfo {author} {\bibfnamefont {Mohamed
  Ezzeldin~A}\ \bibnamefont {Elshaer}}, \bibinfo {author} {\bibfnamefont
  {Florian}\ \bibnamefont {Ettlinger}}, \bibinfo {author} {\bibfnamefont
  {Sunil}\ \bibnamefont {Tatavarty}}, \bibinfo {author} {\bibfnamefont {Marc}\
  \bibnamefont {Bickel}}, \bibinfo {author} {\bibfnamefont {Patrick}\
  \bibnamefont {Bilic}}, \bibinfo {author} {\bibfnamefont {Markus}\
  \bibnamefont {Rempfler}}, \bibinfo {author} {\bibfnamefont {Marco}\
  \bibnamefont {Armbruster}}, \bibinfo {author} {\bibfnamefont {Felix}\
  \bibnamefont {Hofmann}}, \bibinfo {author} {\bibfnamefont {Melvin}\
  \bibnamefont {D’Anastasi}},  \emph {et~al.},\ }\bibfield  {title} {\enquote
  {\bibinfo {title} {Automatic liver and lesion segmentation in {CT} using
  cascaded fully convolutional neural networks and {3D} conditional random
  fields},}\ }in\ \href@noop {} {\emph {\bibinfo {booktitle} {International
  Conference on Medical Image Computing and Computer-Assisted Intervention}}}\
  (\bibinfo {organization} {Springer},\ \bibinfo {year} {2016})\ pp.\ \bibinfo
  {pages} {415--423}\BibitemShut {NoStop}%
\bibitem [{\citenamefont {{\c{C}}i{\c{c}}ek}\ \emph {et~al.}(2016)\citenamefont
  {{\c{C}}i{\c{c}}ek}, \citenamefont {Abdulkadir}, \citenamefont {Lienkamp},
  \citenamefont {Brox},\ and\ \citenamefont {Ronneberger}}]{cciccek20163d}%
  \BibitemOpen
  \bibfield  {author} {\bibinfo {author} {\bibfnamefont {{\"O}zg{\"u}n}\
  \bibnamefont {{\c{C}}i{\c{c}}ek}}, \bibinfo {author} {\bibfnamefont {Ahmed}\
  \bibnamefont {Abdulkadir}}, \bibinfo {author} {\bibfnamefont {Soeren~S}\
  \bibnamefont {Lienkamp}}, \bibinfo {author} {\bibfnamefont {Thomas}\
  \bibnamefont {Brox}}, \ and\ \bibinfo {author} {\bibfnamefont {Olaf}\
  \bibnamefont {Ronneberger}},\ }\bibfield  {title} {\enquote {\bibinfo {title}
  {{3D U-Net}: learning dense volumetric segmentation from sparse
  annotation},}\ }in\ \href@noop {} {\emph {\bibinfo {booktitle} {International
  Conference on Medical Image Computing and Computer-Assisted Intervention}}}\
  (\bibinfo {organization} {Springer},\ \bibinfo {year} {2016})\ pp.\ \bibinfo
  {pages} {424--432}\BibitemShut {NoStop}%
\bibitem [{\citenamefont {Dong}\ \emph {et~al.}(2017)\citenamefont {Dong},
  \citenamefont {Yang}, \citenamefont {Liu}, \citenamefont {Mo},\ and\
  \citenamefont {Guo}}]{dong2017automatic}%
  \BibitemOpen
  \bibfield  {author} {\bibinfo {author} {\bibfnamefont {Hao}\ \bibnamefont
  {Dong}}, \bibinfo {author} {\bibfnamefont {Guang}\ \bibnamefont {Yang}},
  \bibinfo {author} {\bibfnamefont {Fangde}\ \bibnamefont {Liu}}, \bibinfo
  {author} {\bibfnamefont {Yuanhan}\ \bibnamefont {Mo}}, \ and\ \bibinfo
  {author} {\bibfnamefont {Yike}\ \bibnamefont {Guo}},\ }\bibfield  {title}
  {\enquote {\bibinfo {title} {Automatic brain tumor detection and segmentation
  using {U-Net} based fully convolutional networks},}\ }in\ \href@noop {}
  {\emph {\bibinfo {booktitle} {Annual Conference on Medical Image
  Understanding and Analysis}}}\ (\bibinfo {organization} {Springer},\ \bibinfo
  {year} {2017})\ pp.\ \bibinfo {pages} {506--517}\BibitemShut {NoStop}%
\bibitem [{\citenamefont {Sirinukunwattana}\ \emph {et~al.}(2017)\citenamefont
  {Sirinukunwattana}, \citenamefont {Pluim}, \citenamefont {Chen},
  \citenamefont {Qi}, \citenamefont {Heng}, \citenamefont {Guo}, \citenamefont
  {Wang}, \citenamefont {Matuszewski}, \citenamefont {Bruni}, \citenamefont
  {Sanchez} \emph {et~al.}}]{sirinukunwattana2017gland}%
  \BibitemOpen
  \bibfield  {author} {\bibinfo {author} {\bibfnamefont {Korsuk}\ \bibnamefont
  {Sirinukunwattana}}, \bibinfo {author} {\bibfnamefont {Josien~PW}\
  \bibnamefont {Pluim}}, \bibinfo {author} {\bibfnamefont {Hao}\ \bibnamefont
  {Chen}}, \bibinfo {author} {\bibfnamefont {Xiaojuan}\ \bibnamefont {Qi}},
  \bibinfo {author} {\bibfnamefont {Pheng-Ann}\ \bibnamefont {Heng}}, \bibinfo
  {author} {\bibfnamefont {Yun~Bo}\ \bibnamefont {Guo}}, \bibinfo {author}
  {\bibfnamefont {Li~Yang}\ \bibnamefont {Wang}}, \bibinfo {author}
  {\bibfnamefont {Bogdan~J}\ \bibnamefont {Matuszewski}}, \bibinfo {author}
  {\bibfnamefont {Elia}\ \bibnamefont {Bruni}}, \bibinfo {author}
  {\bibfnamefont {Urko}\ \bibnamefont {Sanchez}},  \emph {et~al.},\ }\bibfield
  {title} {\enquote {\bibinfo {title} {Gland segmentation in colon histology
  images: {T}he {GlaS Challenge} contest},}\ }\href@noop {} {\bibfield
  {journal} {\bibinfo  {journal} {Medical image analysis}\ }\textbf {\bibinfo
  {volume} {35}},\ \bibinfo {pages} {489--502} (\bibinfo {year}
  {2017})}\BibitemShut {NoStop}%
\bibitem [{\citenamefont {Zotti}\ \emph {et~al.}(2017)\citenamefont {Zotti},
  \citenamefont {Luo}, \citenamefont {Humbert}, \citenamefont {Lalande},\ and\
  \citenamefont {Jodoin}}]{zotti2017gridnet}%
  \BibitemOpen
  \bibfield  {author} {\bibinfo {author} {\bibfnamefont {Cl{\'e}ment}\
  \bibnamefont {Zotti}}, \bibinfo {author} {\bibfnamefont {Zhiming}\
  \bibnamefont {Luo}}, \bibinfo {author} {\bibfnamefont {Olivier}\ \bibnamefont
  {Humbert}}, \bibinfo {author} {\bibfnamefont {Alain}\ \bibnamefont
  {Lalande}}, \ and\ \bibinfo {author} {\bibfnamefont {Pierre-Marc}\
  \bibnamefont {Jodoin}},\ }\bibfield  {title} {\enquote {\bibinfo {title}
  {{GridNet} with automatic shape prior registration for automatic {MRI}
  cardiac segmentation.}}\ }\href@noop {} {\bibfield  {journal} {\bibinfo
  {journal} {arXiv preprint arXiv:1705.08943}\ } (\bibinfo {year}
  {2017})}\BibitemShut {NoStop}%
\bibitem [{\citenamefont {Holschneider}\ \emph {et~al.}(1990)\citenamefont
  {Holschneider}, \citenamefont {Kronland-Martinet}, \citenamefont {Morlet},\
  and\ \citenamefont {Tchamitchian}}]{holschneider1990real}%
  \BibitemOpen
  \bibfield  {author} {\bibinfo {author} {\bibfnamefont {Matthias}\
  \bibnamefont {Holschneider}}, \bibinfo {author} {\bibfnamefont {Richard}\
  \bibnamefont {Kronland-Martinet}}, \bibinfo {author} {\bibfnamefont {Jean}\
  \bibnamefont {Morlet}}, \ and\ \bibinfo {author} {\bibfnamefont
  {Ph}~\bibnamefont {Tchamitchian}},\ }\bibfield  {title} {\enquote {\bibinfo
  {title} {A real-time algorithm for signal analysis with the help of the
  wavelet transform},}\ }in\ \href@noop {} {\emph {\bibinfo {booktitle}
  {Wavelets}}}\ (\bibinfo  {publisher} {Springer},\ \bibinfo {year} {1990})\
  pp.\ \bibinfo {pages} {286--297}\BibitemShut {NoStop}%
\bibitem [{\citenamefont {Wolterink}\ \emph {et~al.}(2016)\citenamefont
  {Wolterink}, \citenamefont {Leiner}, \citenamefont {Viergever},\ and\
  \citenamefont {I{\v{s}}gum}}]{wolterink2016dilated}%
  \BibitemOpen
  \bibfield  {author} {\bibinfo {author} {\bibfnamefont {Jelmer~M}\
  \bibnamefont {Wolterink}}, \bibinfo {author} {\bibfnamefont {Tim}\
  \bibnamefont {Leiner}}, \bibinfo {author} {\bibfnamefont {Max~A}\
  \bibnamefont {Viergever}}, \ and\ \bibinfo {author} {\bibfnamefont {Ivana}\
  \bibnamefont {I{\v{s}}gum}},\ }\bibfield  {title} {\enquote {\bibinfo {title}
  {Dilated convolutional neural networks for cardiovascular {MR} segmentation
  in congenital heart disease},}\ }in\ \href@noop {} {\emph {\bibinfo
  {booktitle} {Reconstruction, Segmentation, and Analysis of Medical Images}}}\
  (\bibinfo  {publisher} {Springer},\ \bibinfo {year} {2016})\ pp.\ \bibinfo
  {pages} {95--102}\BibitemShut {NoStop}%
\bibitem [{\citenamefont {Wu}\ \emph {et~al.}(2016)\citenamefont {Wu},
  \citenamefont {Shen},\ and\ \citenamefont {Hengel}}]{wu2016high}%
  \BibitemOpen
  \bibfield  {author} {\bibinfo {author} {\bibfnamefont {Zifeng}\ \bibnamefont
  {Wu}}, \bibinfo {author} {\bibfnamefont {Chunhua}\ \bibnamefont {Shen}}, \
  and\ \bibinfo {author} {\bibfnamefont {Anton van~den}\ \bibnamefont
  {Hengel}},\ }\bibfield  {title} {\enquote {\bibinfo {title} {High-performance
  semantic segmentation using very deep fully convolutional networks},}\
  }\href@noop {} {\bibfield  {journal} {\bibinfo  {journal} {arXiv preprint
  arXiv:1604.04339}\ } (\bibinfo {year} {2016})}\BibitemShut {NoStop}%
\bibitem [{\citenamefont {Moeskops}\ \emph {et~al.}(2017)\citenamefont
  {Moeskops}, \citenamefont {Veta}, \citenamefont {Lafarge}, \citenamefont
  {Eppenhof},\ and\ \citenamefont {Pluim}}]{moeskops2017adversarial}%
  \BibitemOpen
  \bibfield  {author} {\bibinfo {author} {\bibfnamefont {Pim}\ \bibnamefont
  {Moeskops}}, \bibinfo {author} {\bibfnamefont {Mitko}\ \bibnamefont {Veta}},
  \bibinfo {author} {\bibfnamefont {Maxime~W}\ \bibnamefont {Lafarge}},
  \bibinfo {author} {\bibfnamefont {Koen~AJ}\ \bibnamefont {Eppenhof}}, \ and\
  \bibinfo {author} {\bibfnamefont {Josien~PW}\ \bibnamefont {Pluim}},\
  }\bibfield  {title} {\enquote {\bibinfo {title} {Adversarial training and
  dilated convolutions for brain {MRI} segmentation},}\ }in\ \href@noop {}
  {\emph {\bibinfo {booktitle} {Deep Learning in Medical Image Analysis and
  Multimodal Learning for Clinical Decision Support}}}\ (\bibinfo  {publisher}
  {Springer},\ \bibinfo {year} {2017})\ pp.\ \bibinfo {pages}
  {56--64}\BibitemShut {NoStop}%
\bibitem [{\citenamefont {Lopez}\ and\ \citenamefont
  {Ventura}(2017)}]{lopez2017dilated}%
  \BibitemOpen
  \bibfield  {author} {\bibinfo {author} {\bibfnamefont {Marc~Moreno}\
  \bibnamefont {Lopez}}\ and\ \bibinfo {author} {\bibfnamefont {Jonathan}\
  \bibnamefont {Ventura}},\ }\bibfield  {title} {\enquote {\bibinfo {title}
  {Dilated convolutions for brain tumor segmentation in {MRI} scans},}\ }in\
  \href@noop {} {\emph {\bibinfo {booktitle} {International MICCAI Brainlesion
  Workshop}}}\ (\bibinfo {organization} {Springer},\ \bibinfo {year} {2017})\
  pp.\ \bibinfo {pages} {253--262}\BibitemShut {NoStop}%
\bibitem [{\citenamefont {Chen}\ \emph {et~al.}(2018)\citenamefont {Chen},
  \citenamefont {Papandreou}, \citenamefont {Kokkinos}, \citenamefont
  {Murphy},\ and\ \citenamefont {Yuille}}]{chen2018deeplab}%
  \BibitemOpen
  \bibfield  {author} {\bibinfo {author} {\bibfnamefont {Liang-Chieh}\
  \bibnamefont {Chen}}, \bibinfo {author} {\bibfnamefont {George}\ \bibnamefont
  {Papandreou}}, \bibinfo {author} {\bibfnamefont {Iasonas}\ \bibnamefont
  {Kokkinos}}, \bibinfo {author} {\bibfnamefont {Kevin}\ \bibnamefont
  {Murphy}}, \ and\ \bibinfo {author} {\bibfnamefont {Alan~L}\ \bibnamefont
  {Yuille}},\ }\bibfield  {title} {\enquote {\bibinfo {title} {{Deeplab}:
  {S}emantic image segmentation with deep convolutional nets, atrous
  convolution, and fully connected {CRF}s},}\ }\href@noop {} {\bibfield
  {journal} {\bibinfo  {journal} {IEEE transactions on pattern analysis and
  machine intelligence}\ }\textbf {\bibinfo {volume} {40}},\ \bibinfo {pages}
  {834--848} (\bibinfo {year} {2018})}\BibitemShut {NoStop}%
\bibitem [{\citenamefont {Anthimopoulos}\ \emph {et~al.}(2018)\citenamefont
  {Anthimopoulos}, \citenamefont {Christodoulidis}, \citenamefont {Ebner},
  \citenamefont {Geiser}, \citenamefont {Christe},\ and\ \citenamefont
  {Mougiakakou}}]{anthimopoulos2018semantic}%
  \BibitemOpen
  \bibfield  {author} {\bibinfo {author} {\bibfnamefont {Marios}\ \bibnamefont
  {Anthimopoulos}}, \bibinfo {author} {\bibfnamefont {Stergios}\ \bibnamefont
  {Christodoulidis}}, \bibinfo {author} {\bibfnamefont {Lukas}\ \bibnamefont
  {Ebner}}, \bibinfo {author} {\bibfnamefont {Thomas}\ \bibnamefont {Geiser}},
  \bibinfo {author} {\bibfnamefont {Andreas}\ \bibnamefont {Christe}}, \ and\
  \bibinfo {author} {\bibfnamefont {Stavroula}\ \bibnamefont {Mougiakakou}},\
  }\bibfield  {title} {\enquote {\bibinfo {title} {Semantic segmentation of
  pathological lung tissue with dilated fully convolutional networks},}\
  }\href@noop {} {\bibfield  {journal} {\bibinfo  {journal} {arXiv preprint
  arXiv:1803.06167}\ } (\bibinfo {year} {2018})}\BibitemShut {NoStop}%
\bibitem [{\citenamefont {Romera}\ \emph {et~al.}(2017)\citenamefont {Romera},
  \citenamefont {Alvarez}, \citenamefont {Bergasa},\ and\ \citenamefont
  {Arroyo}}]{romera2017efficient}%
  \BibitemOpen
  \bibfield  {author} {\bibinfo {author} {\bibfnamefont {Eduardo}\ \bibnamefont
  {Romera}}, \bibinfo {author} {\bibfnamefont {Jos{\'e}~M}\ \bibnamefont
  {Alvarez}}, \bibinfo {author} {\bibfnamefont {Luis~M}\ \bibnamefont
  {Bergasa}}, \ and\ \bibinfo {author} {\bibfnamefont {Roberto}\ \bibnamefont
  {Arroyo}},\ }\bibfield  {title} {\enquote {\bibinfo {title} {Efficient
  convnet for real-time semantic segmentation},}\ }in\ \href@noop {} {\emph
  {\bibinfo {booktitle} {Intelligent Vehicles Symposium (IV), 2017 IEEE}}}\
  (\bibinfo {organization} {IEEE},\ \bibinfo {year} {2017})\ pp.\ \bibinfo
  {pages} {1789--1794}\BibitemShut {NoStop}%
\bibitem [{\citenamefont {Paszke}\ \emph {et~al.}(2016)\citenamefont {Paszke},
  \citenamefont {Chaurasia}, \citenamefont {Kim},\ and\ \citenamefont
  {Culurciello}}]{paszke2016enet}%
  \BibitemOpen
  \bibfield  {author} {\bibinfo {author} {\bibfnamefont {Adam}\ \bibnamefont
  {Paszke}}, \bibinfo {author} {\bibfnamefont {Abhishek}\ \bibnamefont
  {Chaurasia}}, \bibinfo {author} {\bibfnamefont {Sangpil}\ \bibnamefont
  {Kim}}, \ and\ \bibinfo {author} {\bibfnamefont {Eugenio}\ \bibnamefont
  {Culurciello}},\ }\bibfield  {title} {\enquote {\bibinfo {title} {Enet: {A}
  deep neural network architecture for real-time semantic segmentation},}\
  }\href@noop {} {\bibfield  {journal} {\bibinfo  {journal} {arXiv preprint
  arXiv:1606.02147}\ } (\bibinfo {year} {2016})}\BibitemShut {NoStop}%
\bibitem [{\citenamefont {Badrinarayanan}\ \emph {et~al.}(2017)\citenamefont
  {Badrinarayanan}, \citenamefont {Kendall},\ and\ \citenamefont
  {Cipolla}}]{badrinarayanan2017segnet}%
  \BibitemOpen
  \bibfield  {author} {\bibinfo {author} {\bibfnamefont {Vijay}\ \bibnamefont
  {Badrinarayanan}}, \bibinfo {author} {\bibfnamefont {Alex}\ \bibnamefont
  {Kendall}}, \ and\ \bibinfo {author} {\bibfnamefont {Roberto}\ \bibnamefont
  {Cipolla}},\ }\bibfield  {title} {\enquote {\bibinfo {title} {{Segnet}: {A}
  deep convolutional encoder-decoder architecture for image segmentation},}\
  }\href@noop {} {\bibfield  {journal} {\bibinfo  {journal} {IEEE transactions
  on pattern analysis and machine intelligence}\ }\textbf {\bibinfo {volume}
  {39}},\ \bibinfo {pages} {2481--2495} (\bibinfo {year} {2017})}\BibitemShut
  {NoStop}%
\bibitem [{\citenamefont {He}\ \emph {et~al.}(2016)\citenamefont {He},
  \citenamefont {Zhang}, \citenamefont {Ren},\ and\ \citenamefont
  {Sun}}]{he2016deep}%
  \BibitemOpen
  \bibfield  {author} {\bibinfo {author} {\bibfnamefont {Kaiming}\ \bibnamefont
  {He}}, \bibinfo {author} {\bibfnamefont {Xiangyu}\ \bibnamefont {Zhang}},
  \bibinfo {author} {\bibfnamefont {Shaoqing}\ \bibnamefont {Ren}}, \ and\
  \bibinfo {author} {\bibfnamefont {Jian}\ \bibnamefont {Sun}},\ }\bibfield
  {title} {\enquote {\bibinfo {title} {Deep residual learning for image
  recognition},}\ }in\ \href@noop {} {\emph {\bibinfo {booktitle} {Proceedings
  of the IEEE conference on computer vision and pattern recognition}}}\
  (\bibinfo {year} {2016})\ pp.\ \bibinfo {pages} {770--778}\BibitemShut
  {NoStop}%
\bibitem [{\citenamefont {Ioffe}\ and\ \citenamefont
  {Szegedy}(2015)}]{ioffe2015batch}%
  \BibitemOpen
  \bibfield  {author} {\bibinfo {author} {\bibfnamefont {Sergey}\ \bibnamefont
  {Ioffe}}\ and\ \bibinfo {author} {\bibfnamefont {Christian}\ \bibnamefont
  {Szegedy}},\ }\bibfield  {title} {\enquote {\bibinfo {title} {Batch
  normalization: {A}ccelerating deep network training by reducing internal
  covariate shift},}\ }in\ \href@noop {} {\emph {\bibinfo {booktitle}
  {International Conference on Machine Learning}}}\ (\bibinfo {year} {2015})\
  pp.\ \bibinfo {pages} {448--456}\BibitemShut {NoStop}%
\bibitem [{\citenamefont {He}\ \emph {et~al.}(2015)\citenamefont {He},
  \citenamefont {Zhang}, \citenamefont {Ren},\ and\ \citenamefont
  {Sun}}]{he2015delving}%
  \BibitemOpen
  \bibfield  {author} {\bibinfo {author} {\bibfnamefont {Kaiming}\ \bibnamefont
  {He}}, \bibinfo {author} {\bibfnamefont {Xiangyu}\ \bibnamefont {Zhang}},
  \bibinfo {author} {\bibfnamefont {Shaoqing}\ \bibnamefont {Ren}}, \ and\
  \bibinfo {author} {\bibfnamefont {Jian}\ \bibnamefont {Sun}},\ }\bibfield
  {title} {\enquote {\bibinfo {title} {Delving deep into rectifiers:
  {S}urpassing human-level performance on imagenet classification},}\ }in\
  \href@noop {} {\emph {\bibinfo {booktitle} {Proceedings of the IEEE
  international conference on computer vision}}}\ (\bibinfo {year} {2015})\
  pp.\ \bibinfo {pages} {1026--1034}\BibitemShut {NoStop}%
\bibitem [{\citenamefont {Wang}\ \emph {et~al.}(2017)\citenamefont {Wang},
  \citenamefont {Chen}, \citenamefont {Yuan}, \citenamefont {Liu},
  \citenamefont {Huang}, \citenamefont {Hou},\ and\ \citenamefont
  {Cottrell}}]{wang2017understanding}%
  \BibitemOpen
  \bibfield  {author} {\bibinfo {author} {\bibfnamefont {Panqu}\ \bibnamefont
  {Wang}}, \bibinfo {author} {\bibfnamefont {Pengfei}\ \bibnamefont {Chen}},
  \bibinfo {author} {\bibfnamefont {Ye}~\bibnamefont {Yuan}}, \bibinfo {author}
  {\bibfnamefont {Ding}\ \bibnamefont {Liu}}, \bibinfo {author} {\bibfnamefont
  {Zehua}\ \bibnamefont {Huang}}, \bibinfo {author} {\bibfnamefont {Xiaodi}\
  \bibnamefont {Hou}}, \ and\ \bibinfo {author} {\bibfnamefont {Garrison}\
  \bibnamefont {Cottrell}},\ }\bibfield  {title} {\enquote {\bibinfo {title}
  {Understanding convolution for semantic segmentation},}\ }\href@noop {}
  {\bibfield  {journal} {\bibinfo  {journal} {arXiv preprint arXiv:1702.08502}\
  } (\bibinfo {year} {2017})}\BibitemShut {NoStop}%
\bibitem [{\citenamefont {Dice}(1945)}]{dice1945measures}%
  \BibitemOpen
  \bibfield  {author} {\bibinfo {author} {\bibfnamefont {Lee~R}\ \bibnamefont
  {Dice}},\ }\bibfield  {title} {\enquote {\bibinfo {title} {Measures of the
  amount of ecologic association between species},}\ }\href@noop {} {\bibfield
  {journal} {\bibinfo  {journal} {Ecology}\ }\textbf {\bibinfo {volume} {26}},\
  \bibinfo {pages} {297--302} (\bibinfo {year} {1945})}\BibitemShut {NoStop}%
\bibitem [{\citenamefont {Glorot}\ and\ \citenamefont
  {Bengio}(2010)}]{glorot2010understanding}%
  \BibitemOpen
  \bibfield  {author} {\bibinfo {author} {\bibfnamefont {Xavier}\ \bibnamefont
  {Glorot}}\ and\ \bibinfo {author} {\bibfnamefont {Yoshua}\ \bibnamefont
  {Bengio}},\ }\bibfield  {title} {\enquote {\bibinfo {title} {Understanding
  the difficulty of training deep feedforward neural networks},}\ }in\
  \href@noop {} {\emph {\bibinfo {booktitle} {Proceedings of the thirteenth
  international conference on artificial intelligence and statistics}}}\
  (\bibinfo {year} {2010})\ pp.\ \bibinfo {pages} {249--256}\BibitemShut
  {NoStop}%
\bibitem [{\citenamefont {Paszke}\ \emph {et~al.}(2017)\citenamefont {Paszke},
  \citenamefont {Gross}, \citenamefont {Chintala}, \citenamefont {Chanan},
  \citenamefont {Yang}, \citenamefont {DeVito}, \citenamefont {Lin},
  \citenamefont {Desmaison}, \citenamefont {Antiga},\ and\ \citenamefont
  {Lerer}}]{paszke2017automatic}%
  \BibitemOpen
  \bibfield  {author} {\bibinfo {author} {\bibfnamefont {Adam}\ \bibnamefont
  {Paszke}}, \bibinfo {author} {\bibfnamefont {Sam}\ \bibnamefont {Gross}},
  \bibinfo {author} {\bibfnamefont {Soumith}\ \bibnamefont {Chintala}},
  \bibinfo {author} {\bibfnamefont {Gregory}\ \bibnamefont {Chanan}}, \bibinfo
  {author} {\bibfnamefont {Edward}\ \bibnamefont {Yang}}, \bibinfo {author}
  {\bibfnamefont {Zachary}\ \bibnamefont {DeVito}}, \bibinfo {author}
  {\bibfnamefont {Zeming}\ \bibnamefont {Lin}}, \bibinfo {author}
  {\bibfnamefont {Alban}\ \bibnamefont {Desmaison}}, \bibinfo {author}
  {\bibfnamefont {Luca}\ \bibnamefont {Antiga}}, \ and\ \bibinfo {author}
  {\bibfnamefont {Adam}\ \bibnamefont {Lerer}},\ }\bibfield  {title} {\enquote
  {\bibinfo {title} {Automatic differentiation in pytorch},}\ }\href@noop {} {\
   (\bibinfo {year} {2017})}\BibitemShut {NoStop}%
\bibitem [{\citenamefont {Drozdzal}\ \emph {et~al.}(2016)\citenamefont
  {Drozdzal}, \citenamefont {Vorontsov}, \citenamefont {Chartrand},
  \citenamefont {Kadoury},\ and\ \citenamefont {Pal}}]{drozdzal2016importance}%
  \BibitemOpen
  \bibfield  {author} {\bibinfo {author} {\bibfnamefont {Michal}\ \bibnamefont
  {Drozdzal}}, \bibinfo {author} {\bibfnamefont {Eugene}\ \bibnamefont
  {Vorontsov}}, \bibinfo {author} {\bibfnamefont {Gabriel}\ \bibnamefont
  {Chartrand}}, \bibinfo {author} {\bibfnamefont {Samuel}\ \bibnamefont
  {Kadoury}}, \ and\ \bibinfo {author} {\bibfnamefont {Chris}\ \bibnamefont
  {Pal}},\ }\bibfield  {title} {\enquote {\bibinfo {title} {The importance of
  skip connections in biomedical image segmentation},}\ }in\ \href@noop {}
  {\emph {\bibinfo {booktitle} {Deep Learning and Data Labeling for Medical
  Applications}}}\ (\bibinfo  {publisher} {Springer},\ \bibinfo {year} {2016})\
  pp.\ \bibinfo {pages} {179--187}\BibitemShut {NoStop}%
\bibitem [{\citenamefont {Kamnitsas}\ \emph
  {et~al.}(2017{\natexlab{a}})\citenamefont {Kamnitsas}, \citenamefont {Ledig},
  \citenamefont {Newcombe}, \citenamefont {Simpson}, \citenamefont {Kane},
  \citenamefont {Menon}, \citenamefont {Rueckert},\ and\ \citenamefont
  {Glocker}}]{kamnitsas2017efficient}%
  \BibitemOpen
  \bibfield  {author} {\bibinfo {author} {\bibfnamefont {Konstantinos}\
  \bibnamefont {Kamnitsas}}, \bibinfo {author} {\bibfnamefont {Christian}\
  \bibnamefont {Ledig}}, \bibinfo {author} {\bibfnamefont {Virginia~FJ}\
  \bibnamefont {Newcombe}}, \bibinfo {author} {\bibfnamefont {Joanna~P}\
  \bibnamefont {Simpson}}, \bibinfo {author} {\bibfnamefont {Andrew~D}\
  \bibnamefont {Kane}}, \bibinfo {author} {\bibfnamefont {David~K}\
  \bibnamefont {Menon}}, \bibinfo {author} {\bibfnamefont {Daniel}\
  \bibnamefont {Rueckert}}, \ and\ \bibinfo {author} {\bibfnamefont {Ben}\
  \bibnamefont {Glocker}},\ }\bibfield  {title} {\enquote {\bibinfo {title}
  {Efficient multi-scale {3D CNN} with fully connected {CRF} for accurate brain
  lesion segmentation},}\ }\href@noop {} {\bibfield  {journal} {\bibinfo
  {journal} {Medical image analysis}\ }\textbf {\bibinfo {volume} {36}},\
  \bibinfo {pages} {61--78} (\bibinfo {year} {2017}{\natexlab{a}})}\BibitemShut
  {NoStop}%
\bibitem [{\citenamefont {Gorelick}\ \emph {et~al.}(2017)\citenamefont
  {Gorelick}, \citenamefont {Veksler}, \citenamefont {Boykov},\ and\
  \citenamefont {Nieuwenhuis}}]{gorelick2017convexity}%
  \BibitemOpen
  \bibfield  {author} {\bibinfo {author} {\bibfnamefont {Lena}\ \bibnamefont
  {Gorelick}}, \bibinfo {author} {\bibfnamefont {Olga}\ \bibnamefont
  {Veksler}}, \bibinfo {author} {\bibfnamefont {Yuri}\ \bibnamefont {Boykov}},
  \ and\ \bibinfo {author} {\bibfnamefont {Claudia}\ \bibnamefont
  {Nieuwenhuis}},\ }\bibfield  {title} {\enquote {\bibinfo {title} {Convexity
  shape prior for binary segmentation},}\ }\href@noop {} {\bibfield  {journal}
  {\bibinfo  {journal} {IEEE Transactions on Pattern Analysis and Machine
  Intelligence}\ }\textbf {\bibinfo {volume} {39}},\ \bibinfo {pages}
  {258--271} (\bibinfo {year} {2017})}\BibitemShut {NoStop}%
\bibitem [{\citenamefont {Dolz}\ \emph
  {et~al.}(2017{\natexlab{a}})\citenamefont {Dolz}, \citenamefont {Ben~Ayed},\
  and\ \citenamefont {Desrosiers}}]{dolz2017unbiased}%
  \BibitemOpen
  \bibfield  {author} {\bibinfo {author} {\bibfnamefont {Jose}\ \bibnamefont
  {Dolz}}, \bibinfo {author} {\bibfnamefont {Ismail}\ \bibnamefont {Ben~Ayed}},
  \ and\ \bibinfo {author} {\bibfnamefont {Christian}\ \bibnamefont
  {Desrosiers}},\ }\bibfield  {title} {\enquote {\bibinfo {title} {Unbiased
  shape compactness for segmentation},}\ }in\ \href@noop {} {\emph {\bibinfo
  {booktitle} {International Conference on Medical Image Computing and
  Computer-Assisted Intervention}}}\ (\bibinfo {organization} {Springer},\
  \bibinfo {year} {2017})\ pp.\ \bibinfo {pages} {755--763}\BibitemShut
  {NoStop}%
\bibitem [{\citenamefont {Kamnitsas}\ \emph
  {et~al.}(2017{\natexlab{b}})\citenamefont {Kamnitsas}, \citenamefont {Bai},
  \citenamefont {Ferrante}, \citenamefont {McDonagh}, \citenamefont {Sinclair},
  \citenamefont {Pawlowski}, \citenamefont {Rajchl}, \citenamefont {Lee},
  \citenamefont {Kainz}, \citenamefont {Rueckert} \emph
  {et~al.}}]{kamnitsas2017ensembles}%
  \BibitemOpen
  \bibfield  {author} {\bibinfo {author} {\bibfnamefont {Konstantinos}\
  \bibnamefont {Kamnitsas}}, \bibinfo {author} {\bibfnamefont {Wenjia}\
  \bibnamefont {Bai}}, \bibinfo {author} {\bibfnamefont {Enzo}\ \bibnamefont
  {Ferrante}}, \bibinfo {author} {\bibfnamefont {Steven}\ \bibnamefont
  {McDonagh}}, \bibinfo {author} {\bibfnamefont {Matthew}\ \bibnamefont
  {Sinclair}}, \bibinfo {author} {\bibfnamefont {Nick}\ \bibnamefont
  {Pawlowski}}, \bibinfo {author} {\bibfnamefont {Martin}\ \bibnamefont
  {Rajchl}}, \bibinfo {author} {\bibfnamefont {Matthew}\ \bibnamefont {Lee}},
  \bibinfo {author} {\bibfnamefont {Bernhard}\ \bibnamefont {Kainz}}, \bibinfo
  {author} {\bibfnamefont {Daniel}\ \bibnamefont {Rueckert}},  \emph {et~al.},\
  }\bibfield  {title} {\enquote {\bibinfo {title} {Ensembles of multiple models
  and architectures for robust brain tumour segmentation},}\ }in\ \href@noop {}
  {\emph {\bibinfo {booktitle} {International MICCAI Brainlesion Workshop}}}\
  (\bibinfo {organization} {Springer},\ \bibinfo {year} {2017})\ pp.\ \bibinfo
  {pages} {450--462}\BibitemShut {NoStop}%
\bibitem [{\citenamefont {Dolz}\ \emph
  {et~al.}(2017{\natexlab{b}})\citenamefont {Dolz}, \citenamefont {Desrosiers},
  \citenamefont {Wang}, \citenamefont {Yuan}, \citenamefont {Shen},\ and\
  \citenamefont {Ayed}}]{dolz2017deep}%
  \BibitemOpen
  \bibfield  {author} {\bibinfo {author} {\bibfnamefont {Jose}\ \bibnamefont
  {Dolz}}, \bibinfo {author} {\bibfnamefont {Christian}\ \bibnamefont
  {Desrosiers}}, \bibinfo {author} {\bibfnamefont {Li}~\bibnamefont {Wang}},
  \bibinfo {author} {\bibfnamefont {Jing}\ \bibnamefont {Yuan}}, \bibinfo
  {author} {\bibfnamefont {Dinggang}\ \bibnamefont {Shen}}, \ and\ \bibinfo
  {author} {\bibfnamefont {Ismail~Ben}\ \bibnamefont {Ayed}},\ }\bibfield
  {title} {\enquote {\bibinfo {title} {Deep {CNN} ensembles and suggestive
  annotations for infant brain mri segmentation},}\ }\href@noop {} {\bibfield
  {journal} {\bibinfo  {journal} {arXiv preprint arXiv:1712.05319}\ } (\bibinfo
  {year} {2017}{\natexlab{b}})}\BibitemShut {NoStop}%
\bibitem [{\citenamefont {Manj{\'o}n}\ \emph {et~al.}(2018)\citenamefont
  {Manj{\'o}n}, \citenamefont {Coup{\'e}}, \citenamefont {Raniga},
  \citenamefont {Xia}, \citenamefont {Desmond}, \citenamefont {Fripp},\ and\
  \citenamefont {Salvado}}]{manjon2018mri}%
  \BibitemOpen
  \bibfield  {author} {\bibinfo {author} {\bibfnamefont {Jose~V}\ \bibnamefont
  {Manj{\'o}n}}, \bibinfo {author} {\bibfnamefont {Pierrick}\ \bibnamefont
  {Coup{\'e}}}, \bibinfo {author} {\bibfnamefont {Parnesh}\ \bibnamefont
  {Raniga}}, \bibinfo {author} {\bibfnamefont {Ying}\ \bibnamefont {Xia}},
  \bibinfo {author} {\bibfnamefont {Patricia}\ \bibnamefont {Desmond}},
  \bibinfo {author} {\bibfnamefont {Jurgen}\ \bibnamefont {Fripp}}, \ and\
  \bibinfo {author} {\bibfnamefont {Olivier}\ \bibnamefont {Salvado}},\
  }\bibfield  {title} {\enquote {\bibinfo {title} {Mri white matter lesion
  segmentation using an ensemble of neural networks and overcomplete
  patch-based voting},}\ }\href@noop {} {\bibfield  {journal} {\bibinfo
  {journal} {Computerized Medical Imaging and Graphics}\ } (\bibinfo {year}
  {2018})}\BibitemShut {NoStop}%
\bibitem [{\citenamefont {Liu}\ \emph {et~al.}(2017)\citenamefont {Liu},
  \citenamefont {Xu}, \citenamefont {Yin}, \citenamefont {Zhang}, \citenamefont
  {Li},\ and\ \citenamefont {Lu}}]{Liu2017Relationship}%
  \BibitemOpen
  \bibfield  {author} {\bibinfo {author} {\bibfnamefont {Y.}~\bibnamefont
  {Liu}}, \bibinfo {author} {\bibfnamefont {X.}~\bibnamefont {Xu}}, \bibinfo
  {author} {\bibfnamefont {L.}~\bibnamefont {Yin}}, \bibinfo {author}
  {\bibfnamefont {X.}~\bibnamefont {Zhang}}, \bibinfo {author} {\bibfnamefont
  {L.}~\bibnamefont {Li}}, \ and\ \bibinfo {author} {\bibfnamefont
  {H.}~\bibnamefont {Lu}},\ }\bibfield  {title} {\enquote {\bibinfo {title}
  {Relationship between glioblastoma heterogeneity and survival time: {A}n {MR}
  imaging texture analysis},}\ }\href@noop {} {\bibfield  {journal} {\bibinfo
  {journal} {Ajnr Am J Neuroradiol}\ }\textbf {\bibinfo {volume} {38}}
  (\bibinfo {year} {2017})}\BibitemShut {NoStop}%
\end{thebibliography}%

\end{document}